\documentclass[sn-mathphys,Numbered]{sn-jnl}

\usepackage{graphicx}%
\usepackage{multirow}%
\usepackage{amsmath,amssymb,amsfonts}%
\usepackage{amsthm}%
\usepackage{mathrsfs}%
\usepackage[title]{appendix}%
\usepackage{xcolor}%
\usepackage{textcomp}%
\usepackage{manyfoot}%
\usepackage{booktabs}%
\usepackage{algorithm}%
\usepackage{algorithmicx}%
\usepackage{algpseudocode}%
\usepackage{listings}%
\usepackage{subfigure}
\usepackage{threeparttable} 

\theoremstyle{thmstyleone}%
\theoremstyle{thmstyletwo}%

\theoremstyle{thmstylethree}%

\begin{document}

\title[Article Title]{Mastering Autonomous Assembly in Fusion Application with Learning-by-doing: a Peg-in-hole Study}


\author*[1,2,3]{\fnm{Ruochen} \sur{Yin}}\email{ruochen.yin@lut.fi}

\author[1]{\fnm{Huapeng} \sur{Wu}}\email{huapeng.wu@lut.fi}

\author[1]{\fnm{Ming} \sur{Li}}\email{ming.li@lut.fi}
\author[2,4]{\fnm{Yong} \sur{Cheng}}\email{chengyong@ipp.ac.cn}
\author[2]{\fnm{Yuntao} \sur{Song}}\email{songyt@ipp.ac.cn}
\author[1]{\fnm{Heikki} \sur{Handroos}}\email{heikki.handroos@lut.fi}

\affil*[1]{\orgdiv{Laboratory of Intelligent Machines}, \orgname{School of Energy Systems, Lappeenranta University of Technology}, \orgaddress{\street{Yliopistonkatu 34}, \city{ Lappeenranta}, \postcode{53850}, \country{Finland}}}

\affil[2]{\orgdiv{Institute of Plasma Physics Chinese Academy of Sciences (ASIPP)}, \orgname{Chinese Academy of Sciences}, \orgaddress{\city{Hefei}, \postcode{230031}, \country{China}}}

\affil[3]{\orgname{University of Science and Technology of China}, \orgaddress{\city{Hefei}, \postcode{230026}, \country{China}}}

\affil[4]{\orgdiv{Institute of Energy}, \orgname{Hefei Comprehensive National Science Center}, \orgaddress{\city{Hefei}, \postcode{230031}, \country{China}}}


\abstract{Robotic peg-in-hole assembly represents a critical area of investigation in robotic automation. The fusion of reinforcement learning (RL) and deep neural networks (DNNs) has yielded remarkable breakthroughs in this field. However, existing RL-based methods grapple with delivering optimal performance under the unique environmental and mission constraints of fusion applications. As a result, we propose an inventively designed RL-based approach. In contrast to alternative methods, our focus centers on enhancing the DNN architecture rather than the RL model. Our strategy receives and integrates data from the RGB camera and force/torque (F/T) sensor, training the agent to execute the peg-in-hole assembly task in a manner akin to human hand-eye coordination. All training and experimentation unfold within a realistic environment, and empirical outcomes demonstrate that this multi-sensor fusion approach excels in rigid peg-in-hole assembly tasks, surpassing the repeatable accuracy of the robotic arm utilized—0.1 mm—in uncertain and unstable conditions.}




\maketitle

\section{Introduction}\label{sec1}

For decades, researchers worldwide have tirelessly endeavored to cultivate clean, safe, and virtually inexhaustible fusion energy. Over the past decade, the necessity of fusion power as a solution to impending climate change and energy security concerns has become increasingly evident. Presently, the construction of ITER (International Thermonuclear Experimental Reactor) is in progress \cite{iter}, and the conceptual design for DEMO (DEMOstration Power Plant) is under development \cite{demo}. Periodic maintenance is imperative to ensure the functionality of a fusion reactor, necessitating a plethora of remote operations, spanning from the manipulation of sizable structural payloads to the assembly of diminutive bolt components \cite{wilson1983remote}\cite{tada1995remote}\cite{buckingham2016remote}\cite{thomas2013demo}.

The Mascot robot\cite{hamilton2001development}, outfitted with dual arm manipulators, is remotely operated by adept operators to replace diminutive tiles in the "ITER-like Wall"\cite{matthews2011jet}. This human-in-the-loop operation has been employed for JET (Joint European Torus)\cite{jet} for decades and successfully upgraded JET in 2020. However, the drawbacks of such human-in-the-loop manipulation include low efficiency, high skill dependency, a substantial number of well-trained operators, and extended reactor downtime. For instance, it took 17 months for engineers at JET to upgrade the reactor's inner wall and pipes, necessitating the handling of 7,000 assemblies. A brief shutdown period is crucial for the successful operation of the future DEMO reactor. Employing autonomous assembly robots offers a promising solution to achieve a shorter maintenance duration.

Maintenance tasks within the vacuum vessel can be broadly categorized into three primary groups: dismantling, transportation, and installation of heavy-weight components; assembly and transfer of lightweight components; and cutting and welding of pipelines. The tools and methods for transferring the heavy-weight components are currently under discussion, making it premature to automate maintenance studies for the first type of tasks. Consequently, this paper concentrates on lightweight maintenance tasks, which are both numerous and time-consuming. Moreover, the available commercial robotic arm presents an ideal platform for conducting automated maintenance research for such tasks. Among these maintenance tasks, peg-in-hole assembly stands out as the most common and challenging type, thereby warranting focused attention in this paper.

Nonetheless, challenges persist in implementing autonomous robotic solutions within a fusion reactor environment. Firstly, the interior of a fusion reactor's vacuum vessel (VV) constitutes a relatively unstructured setting; both the background and components consist of smooth-surfaced metallic materials. Consequently, the range of applicable optical sensor technology is limited. Moreover, the internal vacuum vessel environment of the fusion reactor undergoes transformation over time due to high temperatures and substantial confinement forces.

To assess the viability of autonomous assembly by robots in fusion applications, we demonstrate a learning-based peg-in-hole assembly task utilizing a standard industrial robot, as it exemplifies the most typical application in assemblies.

\subsection{Task Description}
\begin{figure}[!h]
\centering

\subfigure[]{
\includegraphics[width=5.2cm]{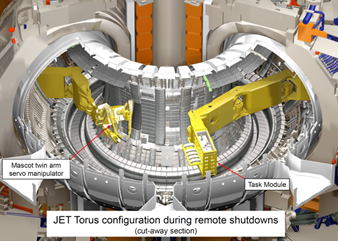}
}
\subfigure[]{

\includegraphics[width=5cm]{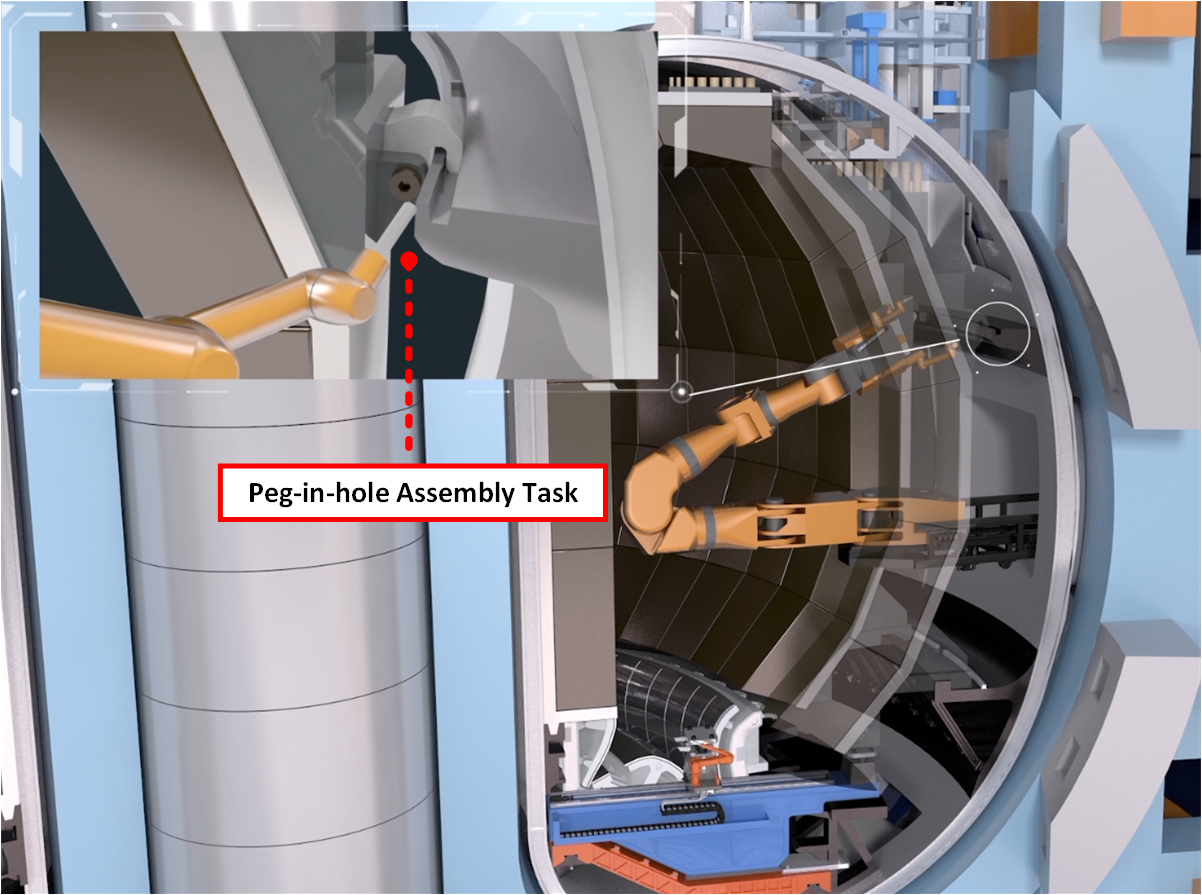}

}
\centering
\caption{The peg-in-hole assembly task in fusion environment: (a) shows the Mascot twin arm system for grasping task and (b) shows the subsequent peg-in-hole assembly task.}
\label{fig:vv1}
\end{figure}

In fusion applications, plasma-facing components necessitate frequent maintenance and replacement. A substantial number of high-precision peg-in-hole assemblies are integral to these tasks. Owing to the unique environment within the VV, the robotic arm must retrieve the peg from the toolbox, situated at a distance from the hole, and subsequently complete the assembly task, as depicted in Figure\ref{fig:vv1}. 

To accommodate an assortment of pegs, a versatile 3-finger-gripper was selected as the end-effector. Concurrently, it was imperative to ensure the connection of the system maintained rigidity throughout the assembly process, signifying the absence of displacement between the peg and the gripper.

Consequently, the peg-in-hole assembly task in this paper is confronted with the following challenges:
\begin{itemize}
\item
In contrast to certain assembly tasks with flexible material items, we are dealing with a rigid one. The tolerance in our application is less than 0.1mm, indicating that vision servo-based approaches could scarcely satisfy the requirement. Simultaneously, the backgrounds, the plasma-facing component, and the peg are all composed of smooth-surfaced metallic materials, resulting in optical sensors, such as RGB-D cameras, exhibiting considerable errors in measuring these polished surfaces.
\item
The precision demanded by the assembly task necessitates a higher degree of repeatability and accuracy than the robotic arm employed.
\item
Owing to inaccuracies in the peg grasping procedure, there exists a degree of randomness in the spatial position relationship between the gripper and the peg.
\item
Since the entire grasping system is highly susceptible to collisions and contact, utmost care must be exercised during the entire assembly process. Thus, forceful collisions and contact are strictly prohibited.
\end{itemize}

\subsection{Related Work}
Peg-in-hole assembly serves as a fundamental aspect of robotic research, representing a prevalent task in maintenance and manufacturing processes. Methodologies encompass vision servo (VS)-based, contact model-based, learning from demonstration, and reinforcement learning (RL)-based approaches.
\subsubsection{VS-based methods}
VS-based methods have long been employed in automated peg-in-hole assembly tasks, encompassing single\cite{jiang2020measurement} or multiple peg assembly. Researchers predominantly utilize traditional computer vision (CV) algorithms, which hinge on extracting the target's geometric features to achieve pose estimation, such as the Circular Hough Transform (CHT)\cite{duda1972use} and Canny algorithm\cite{canny1986computational}. Although a genetic algorithm-based method has been introduced to mitigate calibration error\cite{nagarajan2016vision}, even the employment of an eye-in-hand monocular camera and eye-to-hand stereo camera to establish global and local observation systems respectively \cite{xu2022noncontact}, coupled with markers like chessboards to aid pose estimation\cite{huang2017vision}, these endeavors have only attained millimeter-level accuracy. Consequently, to augment the performance of vision-based pose estimation systems, vision markers have been implemented for both global and local observation systems\cite{jiang2020measurement}. Nevertheless, these VS-based approaches remain constrained by the low resolution of cameras. Simultaneously, during the majority of assembly processes, contact between peg and hole is commonplace; monitoring the contact state to guide the robot in completing the assembly task emerges as a natural concept.

\subsubsection{Contact model-based methods}
Contact model-based methods have been rigorously and extensively investigated in peg-in-hole assembly tasks. Traditional contact model-based strategies hinge on contact model analysis, decomposing the peg-in-hole assembly into two phases: contact-state recognition and compliant control\cite{xu2019compare}. Throughout the assembly task, the system experiences three states: searching, contact, and mated states. As the contact state and its corresponding assembly strategy constitute the most intricate components, contact model-based methods primarily concentrate on this stage.
To expedite the searching stage (where the peg's center is positioned within the clearance region of the hole's center), a spiral searching trajectory\cite{park2020compliant} has been incorporated in some contact model-based methods. Concurrently, several researchers have introduced vision sensors, inspired by vision-based approaches, for initial guidance\cite{zheng2017peg}.
Contact state recognition aims to discern the boundaries of the contact state based on signals from various sensors. For instance, monitoring force and torque alterations due to contact using an F/T sensor is a prevalent method \cite{gai2021feature}\cite{jasim2017contact}. These studies typically differentiate between distinct contact situations through the unique characteristics of force/torque signals, such as one-point, two-point, and three-point contact \cite{tang2016autonomous}. For single peg-in-hole assembly tasks, these three contact situations suffice; however, for certain dual peg-in-hole tasks, the jamming states \cite{zhang2018jamming} or force data analysis \cite{zhang2017force} become more complex, rendering the basic analytical model insufficient. Consequently, contact state recognition relies on statistical methods, such as Hidden Markov Models (HMM)\cite{debus2004contact}, Support Vector Machine (SVM)\cite{jakovljevic2014fuzzy}, and Gaussian Mixtures Model (GMM)\cite{jasim2014contact}\cite{song2021peg}.

The compliant control of contact model-based methods can be divided into two components: decision making and concrete realization. The decision making module is responsible for generating suitable force and moment values for the concrete realization module, according to the geometric constraints\cite{tang2016autonomous}\cite{zhang2018jamming} between the peg and hole. For some sensor-less methods, an additional attractive region constraint may be introduced\cite{su2017study}\cite{su2012sensor}.
Concrete realization can be achieved in several ways. Position control is the most prevalent method—while it boasts high repeatable accuracy, it is primarily utilized for assembly tasks with flexible material items\cite{jasim2014position}. In contrast, force control depends on the force and torque data from the joint, detected by current measurements. For most industrial robots, this data is inaccurate and significantly affects the force control's actual performance. To address environmental uncertainties during the assembly process, some researchers employ hybrid position/force control\cite{de2005sensorless}\cite{park2017compliance}, particularly in rigid assembly tasks.
Contact model-based methods can handle high-precision peg-in-hole assembly tasks in structured environments more effectively than VS-based methods. However, these model-based methods falter in uncertain environments, primarily due to their difficulty in accurately recognizing contact states and their lack of generalizability for varying task environments. Consequently, model-free methods such as learning from demonstration (LFD) and reinforcement learning (RL)-based methods are gaining traction\cite{kober2013reinforcement}.

\subsubsection{LFD methods}

LFD methods hinge on the notion that robots emulate human behavior to accomplish assembly tasks. Typically, robots are taught to execute industrial assembly tasks by delineating key positions and motions using a control device known as a "teach pendant"\cite{b1}. LFD methods encompass three stages: the demonstration phase, the mathematization phase, and the implementation phase. The demonstration phase records human motion trajectories with various sensors, enabling the classification of LFD methods into two categories based on different sensing systems: kinesthetic demonstration\cite{calinon2007learning} and motion capture-based demonstration\cite{lin2014peg}. During kinesthetic demonstration, a human guides the robot, while the robot's built-in joint sensors\cite{paxton2015incremental} record the acceleration, velocity, and position of each joint. Conversely, motion capture-based methods rely on motion capture systems, primarily motion capture cameras\cite{tang2016teach}, occasionally incorporating vision markers\cite{kulic2012incremental} for enhanced accuracy.
Following motion capture, the subsequent stage entails mathematization the motion into a mathematical model. Three prevalent methodologies employed in this phase include hidden Markov models (HMMs)\cite{calinon2010probabilistic}\cite{niekum2013incremental}, Gaussian mixture models (GMMs)\cite{tang2015learning}\cite{kyrarini2019robot}, and dynamic movement primitives (DMPs)\cite{abu2014solving}. DMPs exhibit time-invariance and robustness to spatial perturbations but limited by implicit time dependency. HMMs adeptly model both spatial and temporal motions. Meanwhile, compared with the other two types of methods, GMMs are more robust but prone to overfitting in situations with limited sample data. The reproducing phase is designed to replicate motions once the model has been optimized. LFD methods excel in most single tasks, such as factory assembly lines; but difficult to handle complex, multi-position assembly tasks. Moreover, as online methods, LFD approaches can be time-consuming when deployed in new work environments.
\subsubsection{RL-based methods}

With the rapid advancement of deep learning, RL-based approaches have become the most popular method for peg-in-hole assembly tasks. Unlike other methods, RL-based methods don’t have to design specific strategies for different stages artificially. Instead, they allow the robot to autonomously determine the most appropriate course of action to successfully complete the task.
There are two categories of RL-based methods: value-based methods and policy-based methods\cite{luo2018deep}. value-based methods convert the assembly task into numerous discrete state-action pairs, adjusting the value of each state-action pair during the learning process and optimizing action selection in each state to maximize the total value. When the number of state-action pairs is excessively large, storing and maintaining them becomes time- and space-consuming. Consequently, RL methods combine deep neural networks (DNNs) to parameterize the value functions, such as an value-based method with long short-term memory (LSTM)\cite{b1}. policy-based methods add a scoring network for policy based on value-based methods, such as deep deterministic policy gradient (DDPG)\cite{ren2018learning} methods or model-driven deep deterministic policy gradient (MDDPG)\cite{b2}.

Regarding the sensor system, F/T sensors remain the most popular choice, while some researchers have begun to combine multi-sensors, such as the RGB-D sensor with the F/T sensor\cite{wang2021alignment}. However, this research has only been conducted in a simulation environment. Some researchers have also combined the RGB-D camera with joint information and achieved peg-in-hole assembly tasks in real environments, but this is a flexible assembly task with only centimeter-level accuracy\cite{bogunowicz2020sim2real}.
Since the training of RL-based methods can take millions of steps, it becomes time-consuming in reality due to the limitations of the physical world. As a result, most training processes are carried out in simulation environments\cite{hou2020fuzzy}, and the model is deployed and further optimized in reality after being optimized in simulation. However, overcoming the problem of sim-real gaps\cite{stan2020reinforcement} is not an easy task.

\begin{figure}[!b]
\centering

\subfigure[]{
\includegraphics[width=5cm]{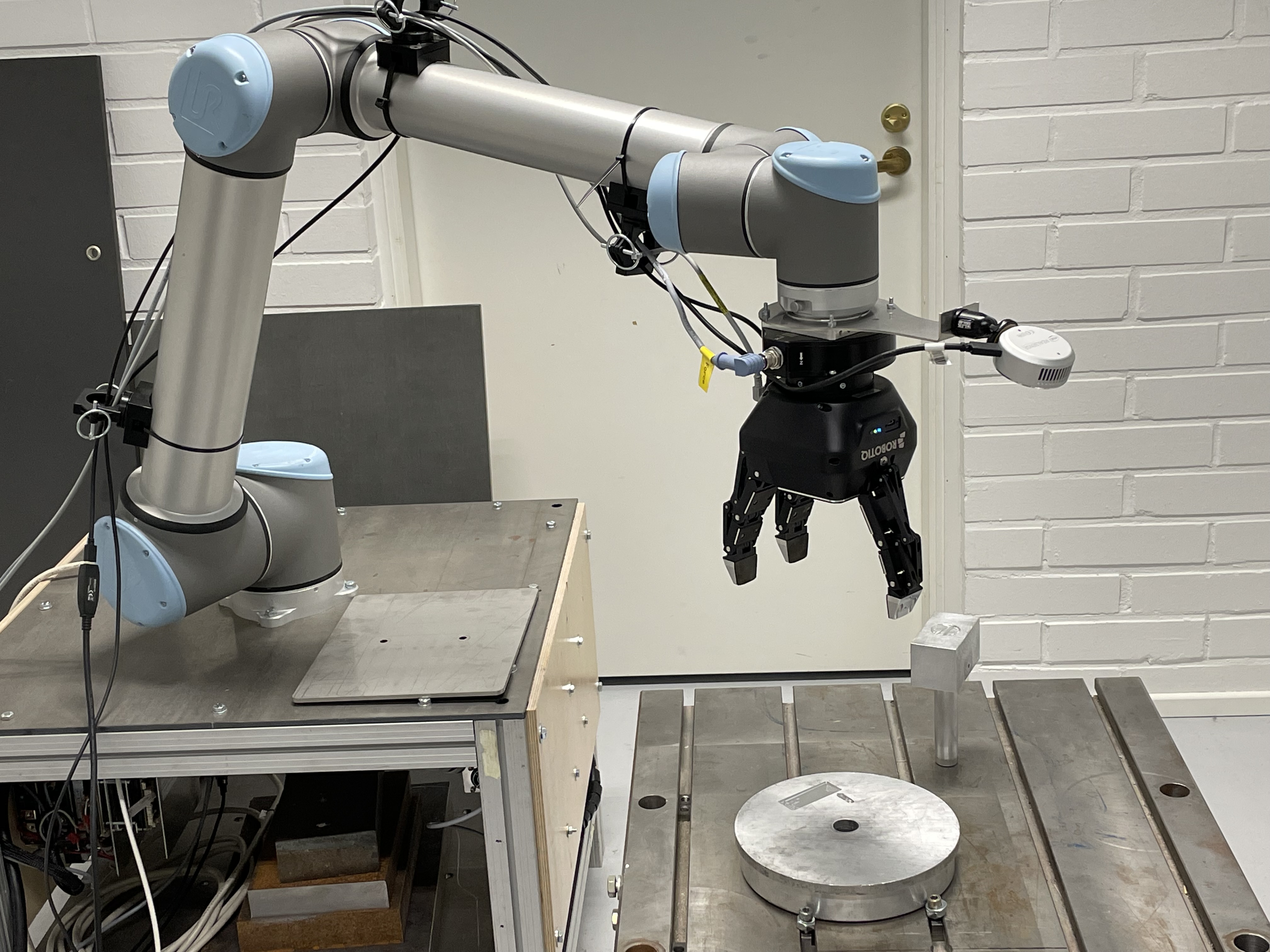}
}
\subfigure[]{

\includegraphics[width=5cm]{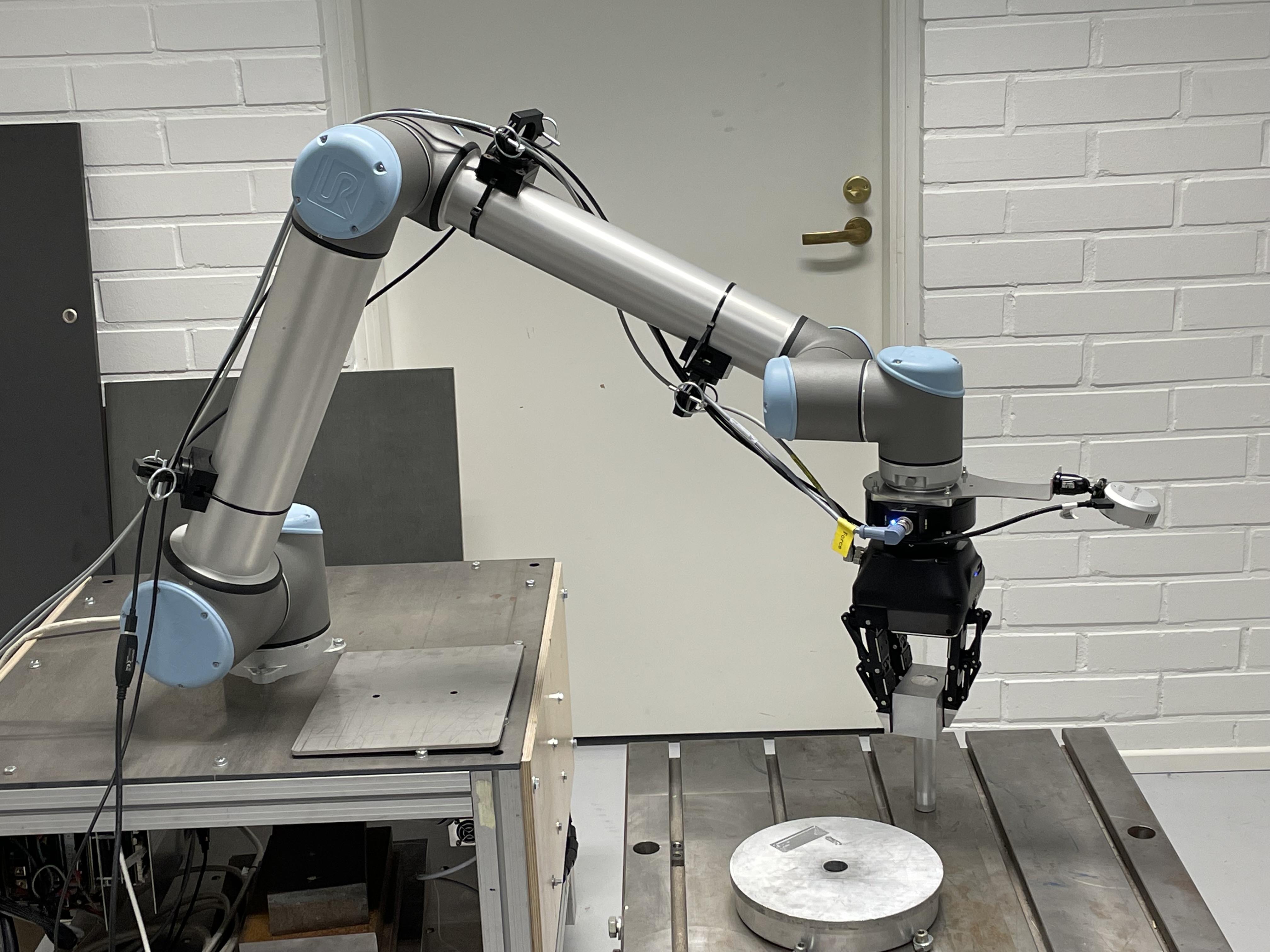}

}
\subfigure[]{
\includegraphics[width=5cm]{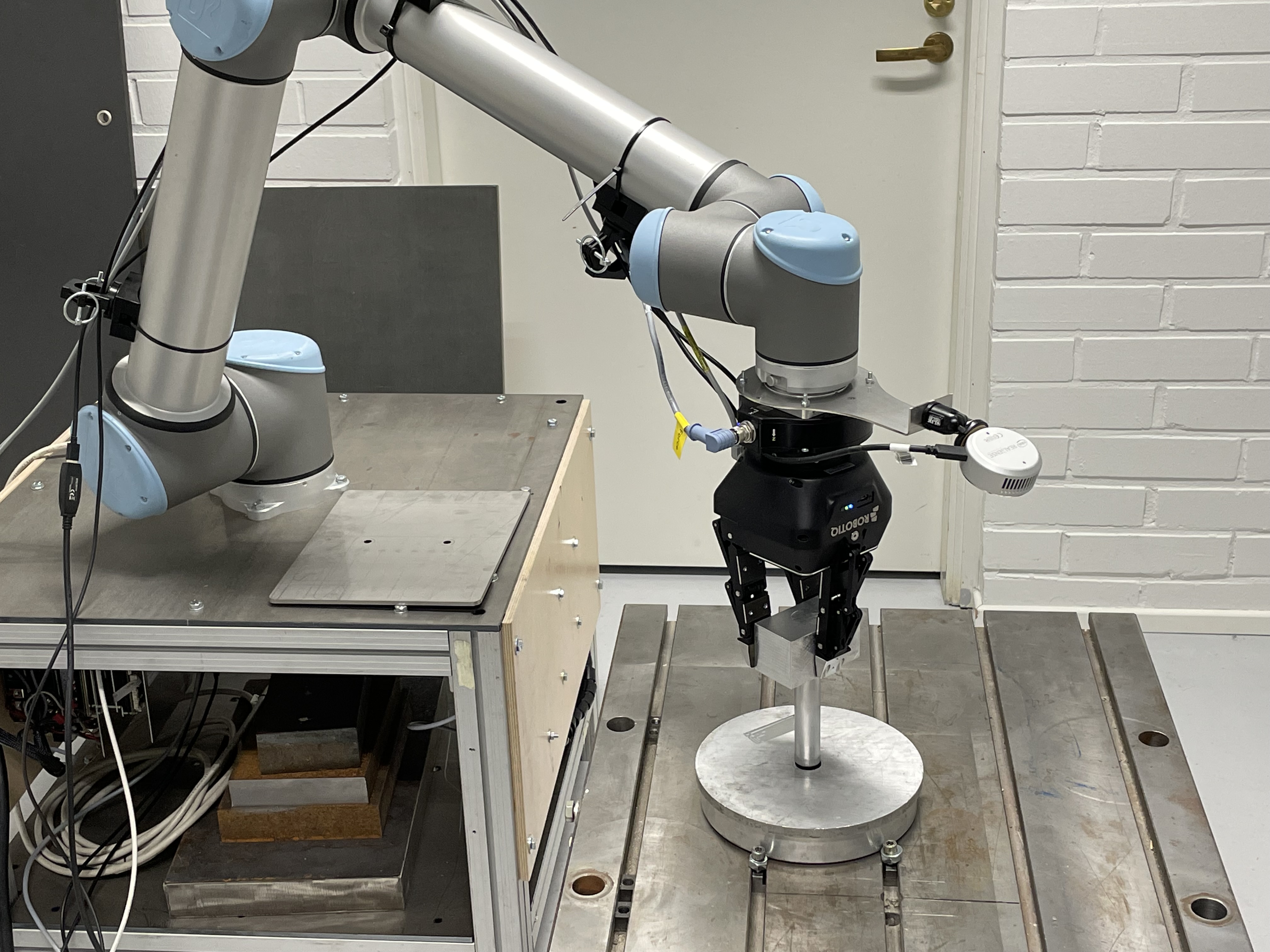}
}
\subfigure[]{

\includegraphics[width=5cm]{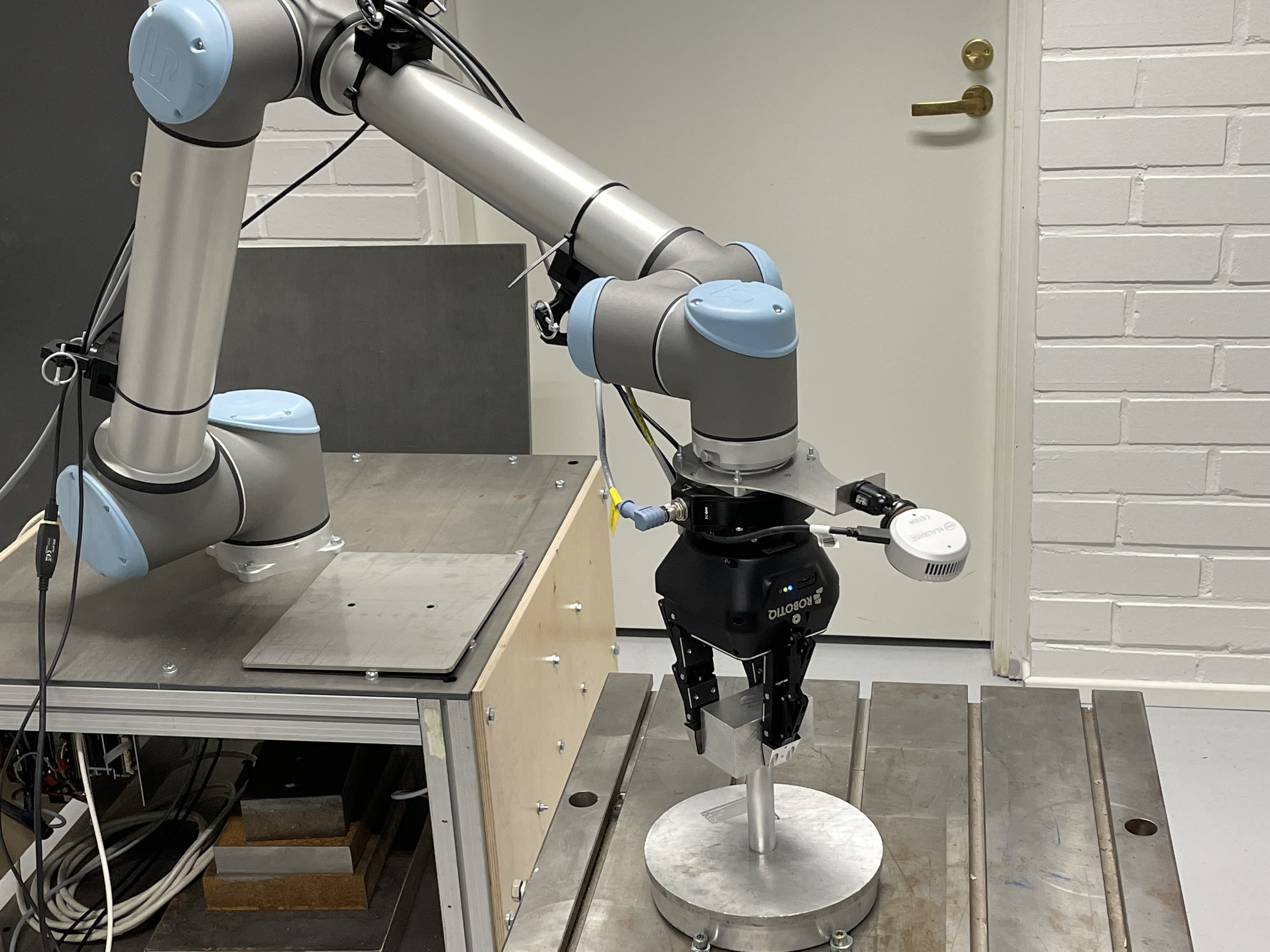}

}
\centering
\caption{The complete  task of the assembly task in Fusion application. Pictures (a) to (d) show the identification phase, the grasping phase, the approaching phase, and the assembly phase.}
\label{fig:vv00}
\end{figure}

\subsection{Contribution}
Our comprehensive task is delineated in Figure \ref{fig:vv00}. To emulate the internal environment of the VV, we incorporate the following configurations in our training and experimental scenarios:
\begin{itemize}
\item
All components and background elements in the training and experiments consist of metallic materials.
\item
The peg-in-hole assembly task commences immediately following the robotic grasping phase and approaching phase, conforming to the actual conditions within the VV.
\end{itemize}

The initial three phases have been accomplished in the preceding paper \cite{Yin2022}. In this paper, we concentrate on the most formidable phase, the assembly phase. Due to the stringent accuracy requirement of less than 0.1 mm for assembling, the vision servo-based method falls short in meeting the precision requirement. Implementing visual markers to enhance accuracy is not feasible within this environment. Additionally, considering that the mounting points are scattered throughout the interior of the VV, the contact model-based and LFD methods are not suitable for handling assembly tasks at multiple locations. To surmount the aforementioned challenges, we introduce a deep reinforcement learning (DRL) approach that amalgamates the RGB image from the camera with the force/torque data from the F/T sensor. Our method bears resemblance to \cite{bogunowicz2020sim2real} and \cite{wang2021alignment}. Nevertheless, we achieve this task in reality rather than solely in a simulated environment, and we confront a rigid assembly task as opposed to a flexible one. Consequently, our task presents an abundance of additional complexities. The method we propose in this paper boasts the following advantages:

\begin{itemize}
\item
We constructed a DRL network capable of accepting both the RGB image from the camera and the force/torque data from the F/T sensor, rendering the method impervious to the initial positional relationship between the peg and the hole. The RGB data facilitates directing the robot arm to expedite alignment of the peg center with the hole center. In essence, it accelerates the search process.
\item
We recognize that the accuracy of the RGB camera alone is insufficient to meet the requirements of this task. Taking inspiration from the way humans utilize hand-eye coordination to successfully execute assembly tasks, we integrated the force/torque data from the F/T sensor with the RGB image, thereby incorporating them as input into our DRL network. These data play important role in different phases.
\item
Owing to the stringent anti-collision requirements and considering control accuracy, we devised a position control strategy with a three-layer cushioning structure (TCS) to ensure the absence of violent collisions or contact throughout the entire process.
\end{itemize}

\begin{figure*}[!t]
    \centering
    \includegraphics[width=13cm]{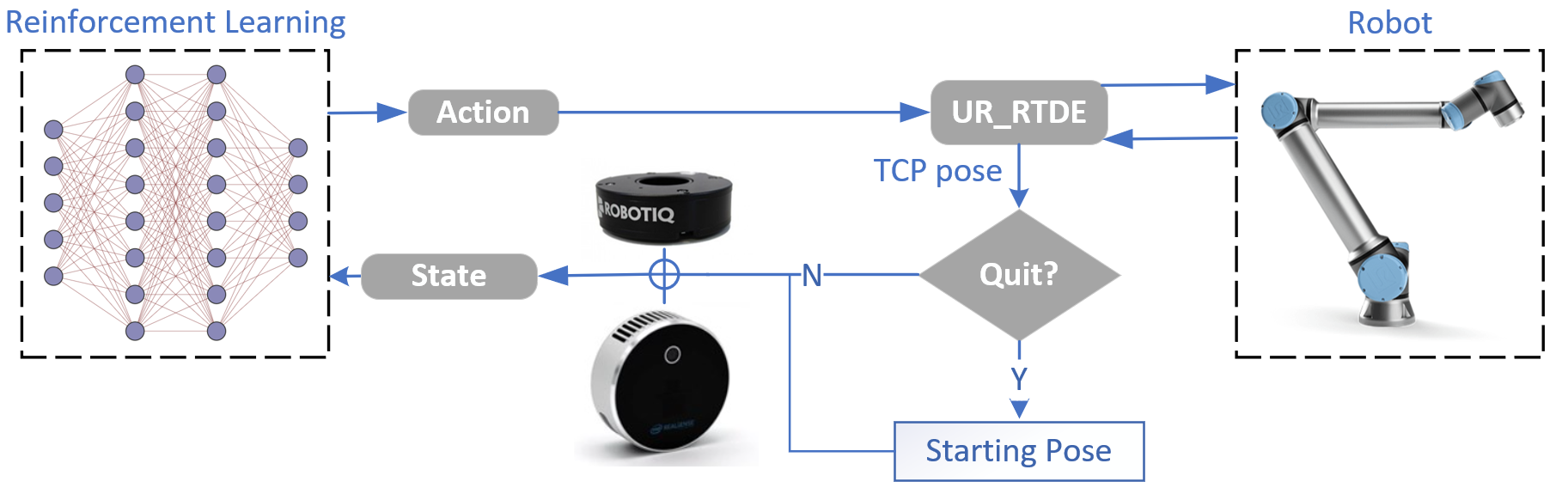}
    \caption{The basic structure of our method.}
    \label{fig:3}
\end{figure*}

\section{Algorithm}
\label{section:2}

The majority of prior research on peg-in-hole assembly tasks relies solely on force data from the F/T sensor. This approach is only viable if the hole center's ground truth position is known. As demonstrated in the work by \cite{b1}, random error within the tolerance range is introduced to the ground truth to simulate positional error. However, in our case, we lack knowledge of the hole center's ground truth position. We can only obtain an approximate result from the RGB-D camera, which exhibits an error exceeding 3mm. Consequently, we integrated the image from an RGB camera into our deep reinforcement learning network and combined it with the force and torque data from the F/T sensor, thereby enabling the robotic arm to complete the task akin to a human worker via a hand-eye coordination mechanism. The architecture of our approach is depicted in Figure \ref{fig:3}, where UR\_RTDE\footnote{\underline{https://sdurobotics.gitlab.io/ur\_rtde/index.html}} is an interface package capable of receiving status information from the robot and transmitting control commands to the UR robot, and TCP represents tool center position. During the assembly process, prior to each movement, we assess the current TCP pose to determine whether we should terminate the current episode (the termination conditions will be elaborated upon later). If the answer is affirmative, the robot reverts to the starting position and transitions to the next training episode; otherwise, the TCP pose is input into the RL network in conjunction with data from the camera and F/T sensor. Subsequently, the RL network generates an action, which is converted into a control signal via the UR\_RTDE module and propels the robot arm. This constitutes one complete step in the cycle.

\subsection{N-step Q Learning}
N-step Q learning (\cite{mnih2016asynchronous}) is a variant of Q learning. This means it is also an value-based method. We chose N-step Q learning as the value-based methods can converge to the global optimal solution and support breakpoint training. Since our training is carried out in a realistic environment throughout, which requires a lot of time, interruptions in the training process are inevitable.
\subsubsection{Problem Statement}
In our scenario, the hole is affixed to the metal table, and the end-effector, a 3-finger gripper, is required to seize the peg from a location on the table and approach the hole; these tasks have been detailed in the preceding paper. The current situation is such that the starting position of the center point at the bottom of the peg (CBP) is approximately 3mm distant, on the X-Y plane, from the hole center, attributable to the data error originating from the RGB-D camera.

In most RL-based peg-in-hole approaches, the task is divided into search and insert phases, with distinct, artificially designed strategies employed to manage these two phases.
\begin{figure}[h]
    \centering
    \includegraphics[width=5cm]{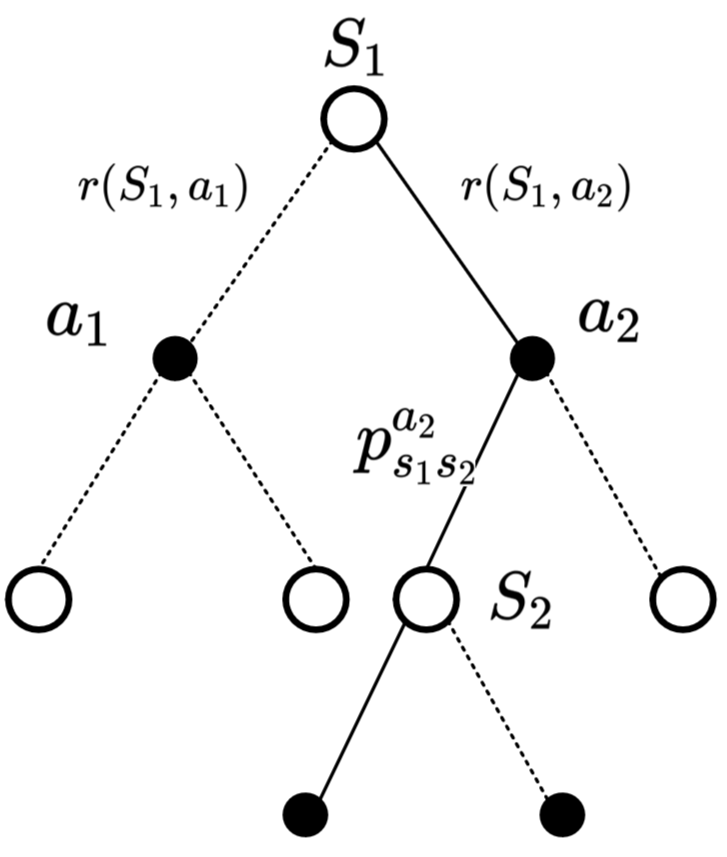}
    \caption{The state transition in reinforcement learning.}
    \label{fig:9_0}
\end{figure}
The objective of this work is to develop an agent that learns to complete the task. This process can be formulated as a Markov decision process (MDP) which could be described as a set of $(S, \mathcal{A}, \mathcal{P}, \mathcal{R}, \gamma)$. $S$ represents the set of state, $\mathcal{A}$ is the action set, $\mathcal{P}$ represents the state transition probability, $\mathcal{R}$ represents the reward set and $\gamma$ is the discount factor. The state transition process could be illustrated as Figure \ref{fig:9_0}. The RL agent observes a state $\mathbf{S}_1$ from the environment, and selects an action $a_2$ based on the Q-value $Q(\mathbf{S}_1,\mathcal{A})$, where $\mathcal{A} = \{a_1,a_2,...,a_n\}$, subsequently receiving reward $r(\mathbf{S}_1,a_2) \in \mathcal{R}$ and transitioning to the next time state $\mathbf{S}_2$ based on the state transition probability $p_{s_1,s_2}^{a_2} \in \mathcal{P}$, then in turn update the $Q(\mathbf{S}_1,a)$ value.

In our scenario, $\mathbf{S}$ is defined as:
\begin{equation}
    \begin{split}
        & \mathbf{P} = \left \{P_x,P_y,P_z,R_x,R_y,R_z\right \} \\
        & \mathbf{F} = \left \{F_x,F_y,F_z, M_x,M_y\right \} \\
        & \mathbf{S} =\left \{\mathbf{I},\mathbf{P},\mathbf{F}\right \}
    \end{split}
\end{equation} 

where $\left \{P_x,P_y,P_z\right \}$ in $\mathbf{P}$ denote the 3D spatial coordinates within the robot arm base coordinate system, and $\left \{R_x,R_y,R_z\right \}$ in $\mathbf{P}$ signify the rotation of the peg relative to the robot arm base coordinate system. $\mathbf{F}$ is the data from the F/T sensor, where $\left \{F_x,F_y,F_z\right \}$ are the forces data and $\left \{M_x,M_y\right \}$ are the moment data; it is essential to note that the F/T sensor is installed close to the end-effector, so the subscript $x,y,z$ in $\mathbf{F}$ represents the axis of the end-effector's coordinate system. The $\mathbf{I}\in \mathbb{R}^{100 \times 100}$ is the cropped grayscale image from the camera, and $\mathbf{S}$ is the set of $\mathbf{I}$, $\mathbf{P}$, and $\mathbf{F}$.

The $\mathbf{a}$ is defined as:
\begin{equation}
    \begin{split}
        & \mathbf{a} =\left \{a_{t}^{x},a_{t}^{y},a_{t}^{z},a_{r}^{x},a_{r}^{y}\right \}
    \end{split}
\end{equation} 

where $a_t$ represents translation actions and $a_r$ denotes rotation actions, with $x,y,z$ in the superscript indicating the axis of the coordinate system. These actions are based on the peg’s coordinate system.

\begin{figure}[!h]
\centering

\subfigure[]{
\includegraphics[width=4cm]{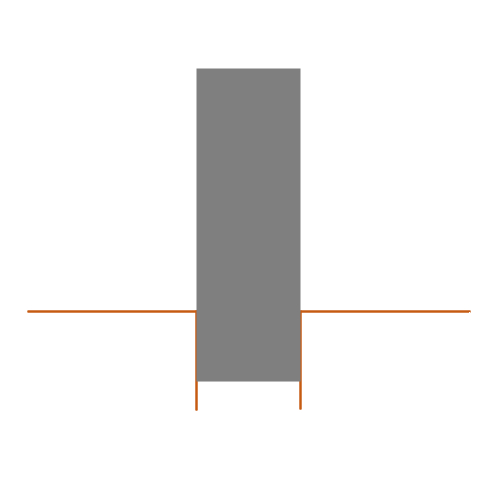}
}
\subfigure[]{

\includegraphics[width=4cm]{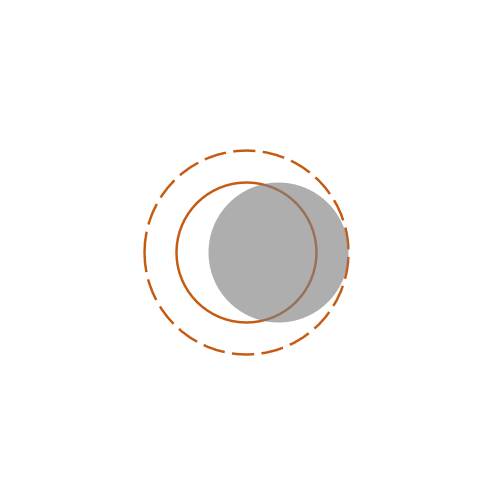}

}
\centering
\caption{The searching region of the peg, (a) is the cross-section of side view, (b) is the top view.}
\label{fig:4}
\end{figure}

\subsubsection{Termination Conditions and Basic Reward Setting}
In this work, we designed several termination conditions as follows:
\begin{itemize}
\item
The robot successfully completes the assembly task; in our case, this means $P_z$ in the current pose should be 1cm lower than the $P_z$ in the starting pose, as shown in Figure~\ref{fig:4}(a).
\item
The distance from CBP to the center point of the hole (CPH) is over 1cm. In this situation, the agent will terminate the current episode and return to the starting pose.
\item
When the agent's moving step surpasses the maximum number $Step_{max}=5000$, we will consider this a failure to complete the task. The agent will end the episode and return to the starting point.
\end{itemize}

For each step, we set the general reward as $r_{gen}= -0.001$. The rationale for assigning a negative general reward is to encourage the robot to complete the task in the fewest steps possible. Concurrently, for each step, we also introduce an extra reward that corresponds to the value of $P_z$ (in meters). Its formula is:
\begin{equation}
    r_{ext} = 75 \times (P_{z}^{t} - P_{z}^{t+1} )
\end{equation} 

where $P_{z}^{t}$ and $P_{z}^{t+1}$ denote the $P_z$ values at the current step and the next step, respectively. Ultimately, if the agent completes the task, it will receive a reward for its success:
\begin{equation}
    r_{suc} = 1 - \frac{Step_f}{Step_{max}} 
\end{equation} 

where $Step_f$ represents the number of steps taken when the agent completes the task.
Simultaneously, to minimize the occurrence of violent contact or collision exceeding a threshold during the assembly process, we incorporate a penalty $r_{pun}$ each time such contact or collision transpires which would be set according to Table \ref{table:2}.
So the final reward $\mathcal{R}$ is:
\begin{equation}
    \mathcal{R}= r_{gen} + r_{ext} + r_{suc} + r_{pun}
\end{equation} 

\begin{figure*}[!t]
    \centering
    \includegraphics[width=13cm]{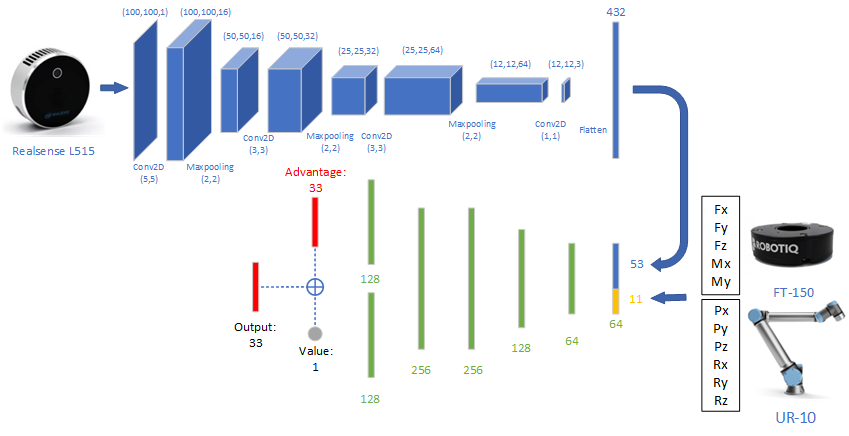}
    \caption{The basic structure of the Multi-Sensor-Based  Dueling Deep Q-learning Network(Dueling-DQN).}
    \label{fig:5}
\end{figure*}

\subsubsection{Search Strategy and Q-value Updating}
At time step $t$, the agent selects an action based on the Q-value of all actions $Q(\mathbf{S}_t,\mathcal{A})$. In our case, we use the $\epsilon-greedy$ strategy.
\begin{equation}
a = \left\{\begin{matrix}
   {\underset{a\in\mathcal{A}}{\mathrm{argmax}}}Q(\mathbf{S}_t,\mathcal{A}) & \mathrm{with\ probalility\ } 1-\epsilon  \\
   \mathrm{randomly\ chose\ } a_k \in \mathcal{A} & \mathrm{with\ probalility\ }\  \epsilon 
\end{matrix}\right. 
\end{equation}

Where $\epsilon$ is a manually set hyperparameter that controls the agent’s exploration. 

The Q-value update follows the equation:
\begin{equation}
\begin{split}
    Q(S_t,a) =
      &Q(S_t,a) + \alpha [\mathcal{R}_{t+1} + \lambda\mathcal{R}_{t+2} + ... \\
      &+\lambda^{k-1}\mathrm{argmax}Q(S_{t+k},\mathcal{A})- Q(S_t,a)]
\end{split}
\label{formula:7}
\end{equation} 

Where $\alpha$ is the learning rate (LR), $\lambda$ is the discount factor, and $k$ is the number of steps that the Q-learning network looks forward.

\subsection{Multi-Sensor-Based Dueling Deep Q-learning Network}
For simple reinforcement learning tasks, all Q-values $Q(\mathbf{S},\mathcal{A})$ can be stored in a Q-table, with reinforcement learning achieved by querying and updating this Q-table. However, when the states become overly complex, maintaining such a vast Q-table is significantly space- and time-consuming. Consequently, a neural network is required, in which the input is a state $\mathbf{S}_t$, and the output is the Q-value of all actions under this state $Q(\mathbf{S}_t,\mathcal{A})$. The neural network provides the same functionality as the Q-table but in a more compact and efficient manner.

In this paper, the structure of the proposed Multi-Sensor-Based Dueling Deep Q-learning Network (MS-Dueling-DQN) is depicted in Figure \ref{fig:5}. The grayscale image $I$ from the camera is input into a CNN branch following several rounds of convolution, and the resulting feature map is flattened into a Fully Connected Layer (FCL). Subsequently, it is concatenated with $\mathbf{P},\mathbf{F}$, originating from the F/T sensor and the UR\_RTDE.

\subsubsection{Calculation of Q-value}
Unlike conventional DQN, Dueling-DQN incorporates an additional branch structure at the output end. Prior to the output layer, the Multi-layer Perceptron (MLP) structure splits into two branches: one referred to as the Advantage Function $Q_{Adv}$ and the other as the Value Function $Q_{Val}$. Meanwhile, the $\oplus$ operator in Figure \ref{fig:5} represents the calculation of the predicted Q-value $Q_{pre}(S_t,\mathcal{A})$:
\begin{equation}
\begin{split}
    Q_{pre}(S_t,\mathcal{A} )= Q_{Val} + Q_{Adv} - Mean(Q_{Adv})
\end{split}
\end{equation} 

where $Mean(Q_{Adv})$ denotes the mean value of $Q_{Adv}$. As designed, in all types of Q-learning, the Q-value represents the value of a state-action pair. However, in Dueling-DQN, the Q-value is regarded as the sum of the value of the state itself $Q_{Val}$ and the value of various actions in this state ($Q_{Adv} - Mean(Q_{Adv})$). The reason for decomposing the final Q value into the $Q_{Val}$  and $Q_{Adv}$ is because sometimes it is unnecessary to know the exact value of each action, so just learning the state value can be enough in some cases. This will speed up the training process.

\begin{figure}[!h]
    \centering
    \includegraphics[width=8cm]{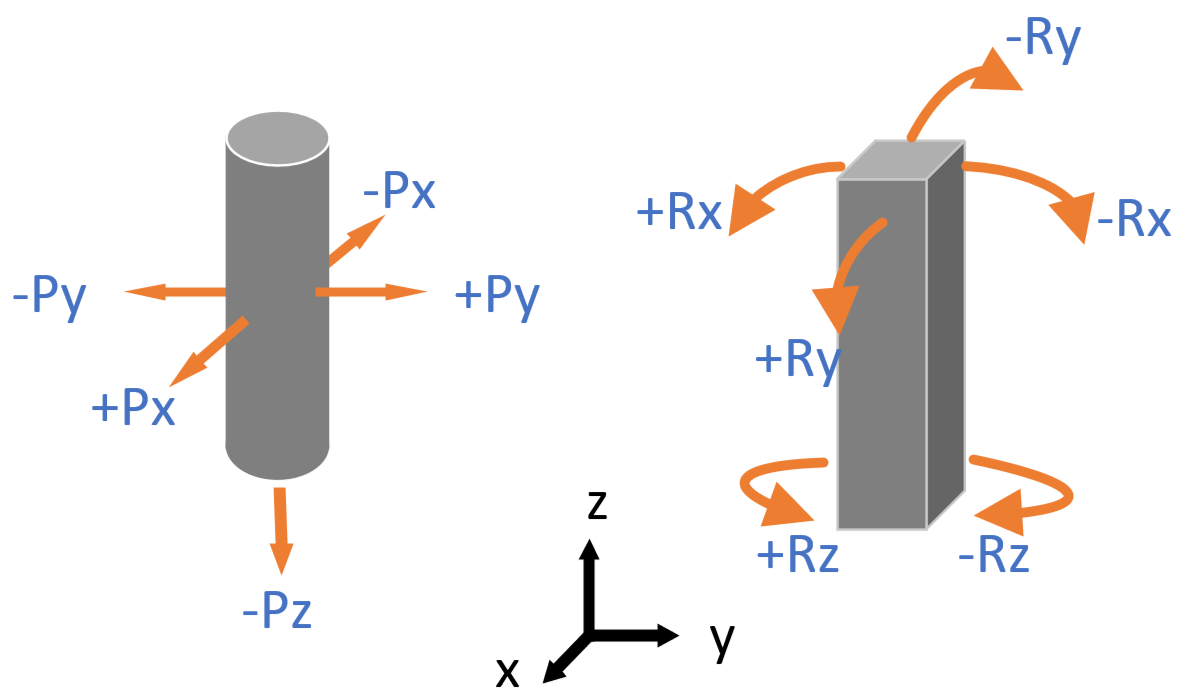}
    \caption{The actions in our case: translation actions and rotation actions. These actions are all referenced to the peg coordinate system, whose origin point is at CBP. And during the rotation actions, the position of CBP is fixed, and the rotation is described in form of Euler Angles.}
    \label{fig:6}
\end{figure}

\subsubsection{Output of the Network}
In a deep reinforcement learning network, each node of the output layer corresponds to an action. In our case, the actions occur in five dimensions: translation actions along axes $x, y, z$, and rotation actions about axes $x, y, z$, as illustrated in Figure \ref{fig:6}. All actions have positive and negative directions, except for the translation action on the z-axis. Additionally, we designed three different step lengths for each action. For the translation actions, the step lengths are ${0.1, 0.5, 1}$ in millimeters, and for the rotation actions, the step lengths are ${0.05,0.08,0.1}$ in degrees. Consequently, there are 33 nodes in the output layer, as depicted in Figure \ref{fig:5}.

It is worth noting that we opted for a position control strategy with cushioning for this task; this is why the translation actions are represented by ${P_x, P_y, P_z }$ in Figure \ref{fig:6}. The control strategy is comprehensively explained in the following section.

\subsubsection{Loss Function and Optimizer}
The loss function is defined as:
\begin{equation}
\begin{split}
    & L = \frac{1}{2} {(Q_{tar}(S_t,a) - Q_{pre}(S_t,a) )} ^{2} \\
    & Q_{pre}(S_t,a) =  {\mathrm{argmax}}Q_{pre}(S_t,\mathcal{A} )
\end{split}
\label{formula:9}
\end{equation} 

where $Q_{pre}(S_t,\mathcal{A})$ is outputted by the network, $Q_{tar}(S_t,a)$ could be calculated as:
\begin{equation}
\begin{split}
    Q_{tar}(S_t,a) &=  Q_{pre}(S_t,a) + \alpha (\mathcal{R}_{t+1} + \lambda\mathcal{R}_{t+2} + ... \\
      & +\lambda^{k-1}\mathrm{argmax}Q_{pre}(S_{t+k},\mathcal{A})- Q_{pre}(S_t,a) )
\end{split}
\label{formula:10}
\end{equation} 

where the $Q_{pre}(S_{t+k},\mathcal{A})$ is also outputted by the network just like $Q_{pre}(S_t,\mathcal{A})$. 

In the deep reinforcement learning (DRL) network, training aims to optimize the network to better fit Formula \ref{formula:7}. In this work, the chosen optimizer for the network is Nesterov-accelerated Adaptive Moment Estimation (Nadam) \cite{Nadam}.Nadam is a popular optimization algorithm that combines the benefits of both Nesterov-accelerated gradient (NAG) \cite{botev2017nesterov} and Adam \cite{kingma2014adam} optimization algorithms. NAG is known for its faster convergence rate, while the Adam optimizer provides an adaptive learning rate. The combination of these two techniques results in a more efficient and robust optimization algorithm that can handle a wide range of optimization problems.

\subsubsection{Parameter Setting}
Among the numerous parameters, the parameter $\epsilon$ within the $\epsilon$-greedy strategy remains one of the most crucial and intriguing. The value of $\epsilon$ diminishes in accordance with the training steps, adhering to the subsequent formula.
\begin{equation}
\begin{split}
    \epsilon = \epsilon_{min} + (\epsilon_{max}-\epsilon_{min})(1-\frac{Steps_{accu}}{Indx_{decay}}) 
\end{split}
\end{equation} 

Wherein, $\epsilon_{min}=0.1$, $\epsilon_{max}=1$, $Indx_{decay}=30000$, and $Steps_{accu}$ represents the aggregate number of motion steps throughout all episodes during the ongoing training. As inferred from the formula, when $\epsilon$ approaches 1, the robot's movements are predominantly random. This arises due to the limited exploration of the environment and the minimal state experiences during the initial stages of training, rendering the acquired knowledge unreliable. Consequently, augmenting the randomness of movement enables the robot to investigate a greater number of unfamiliar states. Nevertheless, since training and experimentation transpire within a real environment and consume considerable time, occasional interruptions are inescapable. In such instances, $Steps_{accu}$ reverts to 0. Thus, after adequate training and satisfactory performance, $\epsilon$ is assigned a fixed small value.

Some other important parameters are shown in the table blow.

\subsection{Control Strategy}
This subsection delves into the intricacies of the position control strategy strategy and the tri-layer cushioning architecture devised to avert forceful collisions or contact in this undertaking.

\subsubsection{Position Control Strategy}
As previously mentioned, the robot arm's TCP pose can be precisely regulated in 3D Cartesian space using the UR\_RTDE package. To ensure a consistent CBP during rotational maneuvers, it is imperative to determine the precise TCP pose.

Per the design, the angle alters exclusively in one dimension throughout each rotation action. Presuming the rotation action involves a rotation of $\Delta\gamma$ degrees about the X-axis, given that the rotation sequence adheres to roll-pitch-yaw (R-P-Y) angles, the TCP $P_{new}$ can be computed as follows:

\begin{equation}
\begin{split}
&M_{old}  = R_{z}(\alpha)R_{y}(\beta)R_{x}(\gamma) \\
        & = \begin{bmatrix}
 C_{\alpha}  & -S_{\alpha}  & 0 \\
 S_{\alpha} & C_{\alpha}  & 0 \\
 0 & 0 & 1
\end{bmatrix}
 \begin{bmatrix}
 C_{\beta}   & 0  & S_{\beta}   \\
 0 & 1  & 0 \\
 -S_{\beta} & 0 & C_{\beta}
\end{bmatrix}
 \begin{bmatrix}
 1   & 0  & 0  \\
 0 & C_{\gamma } & -S_{\gamma } \\
 0 & S_{\gamma } & C_{\gamma } 
\end{bmatrix} \\
\\
&M_{new} = R_{z}(\alpha)R_{y}(\beta)R_{x}(\gamma +\Delta\gamma) \\
        & = \begin{bmatrix}
 C_{\alpha}  & -S_{\alpha}  & 0 \\
 S_{\alpha} & C_{\alpha}  & 0 \\
 0 & 0 & 1
\end{bmatrix}
 \begin{bmatrix}
 C_{\beta}   & 0  & S_{\beta}   \\
 0 & 1  & 0 \\
 -S_{\beta} & 0 & C_{\beta}
\end{bmatrix}
 \begin{bmatrix}
 1   & 0  & 0  \\
 0 & C_{\gamma +\Delta\gamma} & -S_{\gamma +\Delta\gamma} \\
 0 & S_{\gamma +\Delta\gamma} & C_{\gamma +\Delta\gamma} 
\end{bmatrix} \\
\\
&P_{new} = M_{old}{\begin{bmatrix}
 0 & 0 & z_{peg}
\end{bmatrix}
} ^ T + P_{old} - M_{new}{\begin{bmatrix}
 0 & 0 & z_{peg}
\end{bmatrix}
}^ T
\end{split}
\end{equation} 

Wherein, $\gamma,\beta,\alpha$ represent the R-P-Y angles, and $P_{old}$ denotes the initial TCP, all of which can be readily obtained from the UR\_RTDE. $C_x, S_x$ serve as abbreviations for $cos(x), sin(x)$ and analogous symbols hold for the other variables. $z_{peg}$ signifies the Z-axis coordinates of the CBP within the end-effector coordinate system.

Translation actions, likewise, are in reference to the peg coordinate system and can be effortlessly executed through forward kinematics.

At the heart of our control strategy lies the prevention of forceful contact and collisions. Consequently, we incorporated an F/T sensor between the robot arm's final wrist and the end-effector, enabling us to monitor the feedback force and torque of contact throughout each action. Simultaneously, we devised a tri-layer cushioning structure (TCS) to address various contact scenarios.

\begin{table}[!h]
\caption{Some important parameter in our algorithm}\label{table:1}

\begin{tabular}{ccl}
\bottomrule
Symbol & Value & \multicolumn{1}{c}{Description} \\ \midrule
    $\lambda$        &      $0.95  $         &  The discount factor.                \\  
    $\alpha$        &       $1\times10^{-4}$        &      The learning rate.       \\ 
    $k$      &    $ 4 $           & The looking forward steps.           \\ 
    $Step_{max}$       &      $5000 $        & The maximum number of steps in each episode. \\ 
    Batch Size       &      $32$         &  Batch Size during the training.   \\ 
    Buffer Size       &      $1\times10^{5}$         &  The size of the buffer zone.   \\ 
    Update Rate      &      $1\times10^{3}$    & The number of the steps between the target network's updating.                    \\ \bottomrule
\end{tabular}

\end{table}

\subsubsection{Three-layer Cushioning Structure}
\begin{figure}[!h]
    \centering
    \includegraphics[width=8cm]{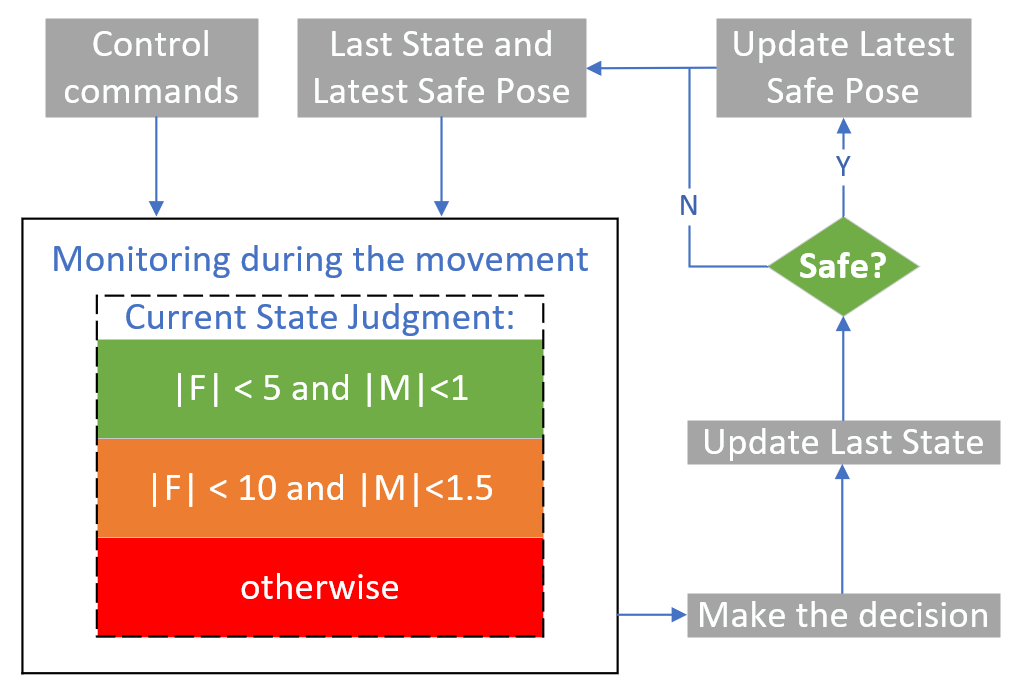}
    \caption{The structure of TCS, where F and M indicate the force and moment (torque). The green zone is the safe state, the orange zone is the warning state, and the red zone is the dangerous state. The TCS would monitor the force and torque data and make the decision in real-time to avoid violent contact and collisions from happening. After each movement, TCS will update the \textbf{Last State} and \textbf{Latest Safe Pose}.}
    \label{fig:7}
\end{figure}

The fundamental structure of TCS is depicted in Figure \ref{fig:7}. TCS continually monitors data from the F/T sensors during each movement of the robot arm; relying on the information gathered from the F/T sensor, the robot arm's current state is categorized as safe, warning, or dangerous. Concurrently, we preserve two global variables, namely \textbf{Last State} and \textbf{Latest Safe Pose}. Guided by these states, the TCS adheres to the decision-making logic presented in Table \ref{table:2} to circumvent forceful contact and collisions.

\begin{table}[!h]
\caption{The decision logic of TCS}\label{table:2}

\begin{tabular}{cclc}
\bottomrule
Last State & Current State & \multicolumn{1}{c}{Decision} & Punishing Reward $r_{pun}$ \\ \midrule
    {\color{green}Safe }        &      {\color{green}Safe }          &  Do nothing.            &       0     \\  
    {\color{green}Safe }        &      {\color{orange}Waring }         &    \begin{tabular}[c]{@{}l@{}}Stop the movement \\ and stay at current pose.\end{tabular}     &           0          \\ 
    {\color{orange}Waring }       &     {\color{green}Safe }            &  Do nothing.      &        0          \\ 
    {\color{orange}Waring }       &      {\color{orange}Waring }         & Do nothing.         &       0          \\ 
    {\color{orange}Waring }       &      {\color{red}Dangerous }         &  
    \begin{tabular}[c]{@{}l@{}}Cancel the movement and move \\ back to the \textbf{Latest Safe Pose}.\end{tabular}
                &      -0.003      \\ 
    {\color{red}Dangerous }        &      {\color{red}Dangerous }    & 
     \begin{tabular}[c]{@{}l@{}}When this appears ten times in a row, \\ quit this episode.\tnote{1} \end{tabular} 
                 &       -0.01     \\ \bottomrule

\end{tabular}
\begin{tablenotes}    
        \footnotesize               
        \item[1] This constitutes an infrequent yet existent scenario, wherein the robot becomes ensnared in an unusual pose and cannot extricate itself through alternative TCS measures. In such instances, the task is deemed a failure. The robot subsequently aborts the current episode, reverts to the starting position, and commences the subsequent episode..          
      \end{tablenotes}  
\end{table}

\section{Training and Experiment}
Contrary to several DRL tasks, we conduct the training and experimentation on the agent in a real-world setting rather than a simulated environment. This approach circumvents the sim-real gap issue, albeit at the expense of being time-consuming. 
\begin{figure}[!h]
    \centering
    \includegraphics[width=8cm]{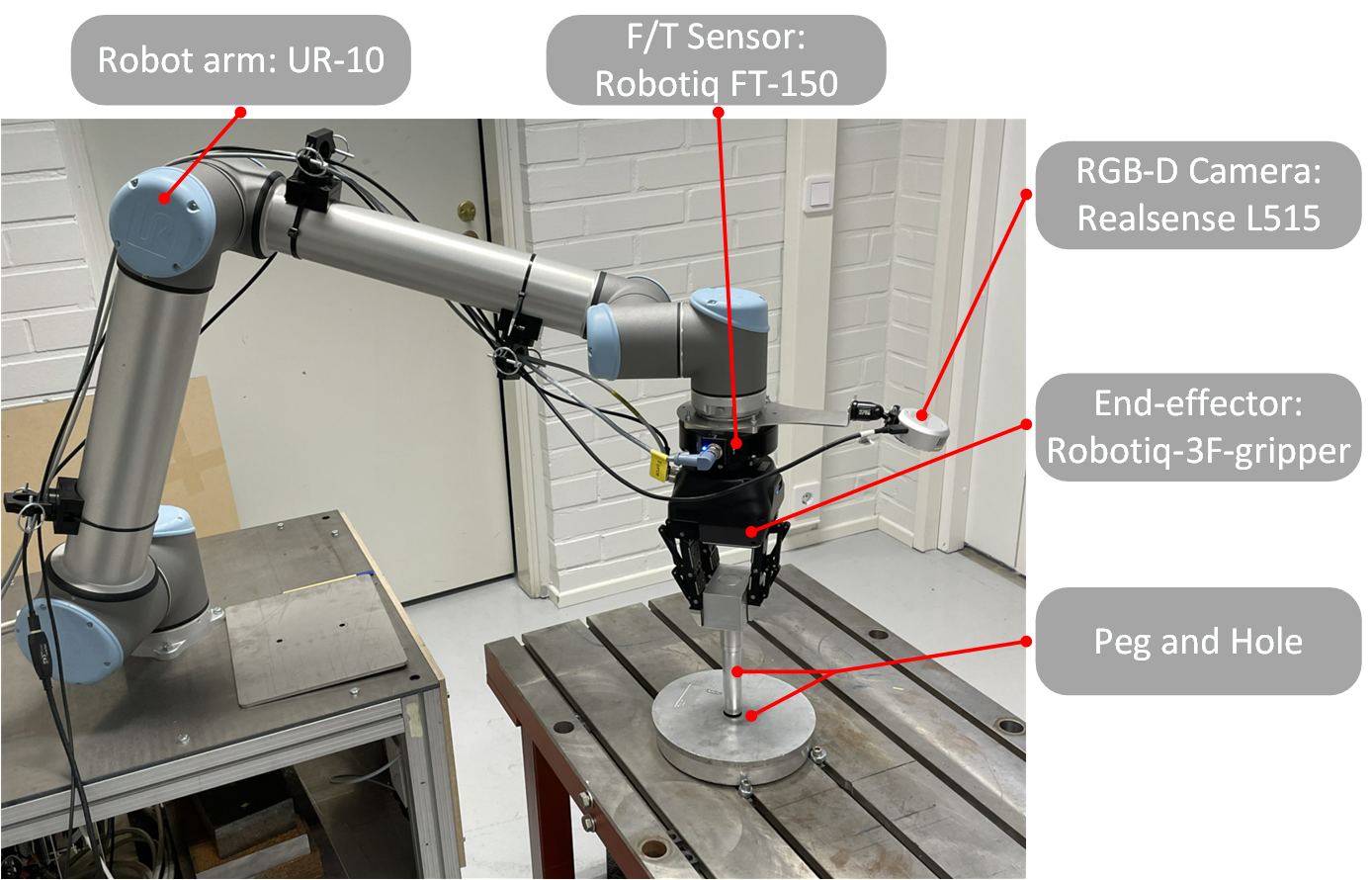}
    \caption{The platform in our task.}
    \label{fig:8}
\end{figure}

\subsection{Experiment platform}
Our experimental platform comprises a host computer (HC), an array of sensors, and a robot arm. Training and experimentation transpire on an HC equipped with an Nvidia RTX 2080Ti GPU, Intel i7-10700k CPU, and 64GB of RAM. The robot arm and various sensors are illustrated in Figure \ref{fig:8}. Communication between the robot arm, end-effector, and HC occurs via the TCP/IP protocol, while the F/T sensor is connected to the HC through a serial port, and the RGB-D camera interfaces with the HC through a USB3.0 port.

\subsection{Training process}
Training for this task spanned over 2000 hours, underscoring the significance of the method's versatility. During the training phase, we meticulously fine-tuned the DRL model's various hyperparameters and standardized the input data. In this subsection, we revisit the salient aspects of the training process that facilitated enhancements in the model's performance and versatility.

\subsubsection{Data Pre-processing}
We will restore the data from F/T sensor at the beginning of the training as $\mathbf{F}_{star}$. Then all the input F/T data $\mathbf{F}_{nor}$ are calculated as:
\begin{equation}
\begin{split}
    \mathbf{F}_{nor} = \mathbf{F} - \mathbf{F}_{star}
\end{split}
\label{formula:12}
\end{equation} 

where $\mathbf{F}$ is the raw data from the F/T sensor. 

As depicted in Figure 2 (d), an eye-in-hand camera was employed for this task, with the positional arrangement between the peg and the robot arm being constant. Consequently, we were able to identify four predetermined points in the image and utilize the resulting rectangular area as the region of interest (ROI). Subsequently, the RGB image was cropped, converted to grayscale, and normalized as follows:
\begin{equation}
\begin{split}
    I_{nor}(u,v) =  1 - \frac{I_{gray}(u,v)}{256} 
\end{split}
\label{formula:13}
\end{equation}

Furthermore, the camera we selected is an RGB-D camera capable of providing 3D data. In our previous publication \cite{Yin2022}, we presented an instance segmentation technique that allowed us to obtain the approximate location of the CPH denoted as $\mathbf{P}_{hc}= \{P^{'}_x,P^{'}_y,P^{'}_z\}$, which has an error margin of approximately 3 to 4 mm. As a result, we can compute the input pose data $\mathbf{P}_{nor}$ as follows:
\begin{equation}
\begin{split}
    & \mathbf{P}_t = \{P_x, P_y, P_z\} - \{P^{'}_x,P^{'}_y,P^{'}_z\} \\
    & \mathbf{P}_r = \{R_x, R_y, R_z\}   \\
    & \mathbf{P}_{nor} = \{\mathbf{P}_t, \mathbf{P}_r\}
\end{split}
\label{formula:14}
\end{equation} 

where $P_x, P_y, P_z,R_x, R_y, R_z$ are the data get from \\ UR\_RTDE.

\subsubsection{From One-step to N-step}
Despite initially opting for N-step Q-learning, we conducted a performance comparison between one-step and N-step Q-learning approaches for this particular task.

\begin{figure}[!h]
\centering

\subfigure[]{
\includegraphics[width=5cm]{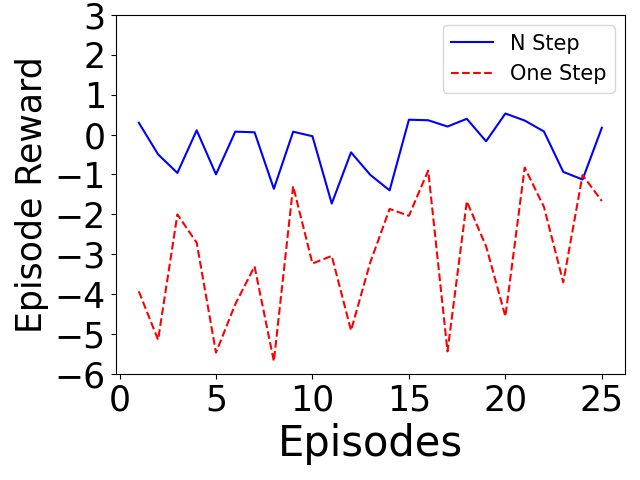}
}
\subfigure[]{

\includegraphics[width=5cm]{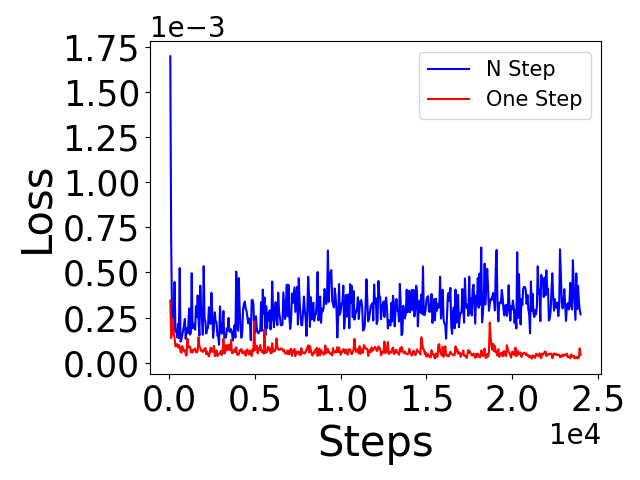}

}
\centering
\caption{The episode reward (ER) and loss comparison between the N-step and one step DRL models. Please note that the data in (a) is not recorded from the very beginning of training but when the model is relatively stable after some training.}
\label{fig:9}
\end{figure}

Figure \ref{fig:9} indicates that, in comparison to the N-step model, the one-step model exhibits lower loss and episode reward during training. This suggests that the one-step model converges more easily. However, the N-step model demonstrates superior performance during the task execution. Essentially, the N-step model has a higher upper bound for the best model, but it takes longer for the network to converge to this optimal state.

Clearly, when the network is capable of converging effectively to the ideal model, the upper limit of the model becomes more important. However, employing a large step number can hinder the network's ability to fit the ideal model and adversely affect the final performance. Therefore, the chosen step number should strike a balance, neither being too large nor too small. After thorough comparison, we settled on a looking forward steps number of four, which yielded peak performance for the task.

\subsubsection{Reward Setting}
The selection of an appropriate reward $\mathcal{R}$ is crucial for the successful design of a reinforcement learning model. Researchers can influence the agent's behavior by assigning different reward values.

In our case, the general reward $r_{gen}$ is set as a negative value, motivating the agent to complete the task as quickly as possible. As for the punishing reward $r_{pun}$,If the punishment is too lenient, the robot may exhibit reckless behavior and frequently engage in collisions and contact. On the other hand, if the punishment is too severe, the robot may become overly cautious or even avoid any contact altogether, making it impossible to complete the assembly task.

To strike a balance between efficiency and safety, significant effort was dedicated to fine-tuning $r_{pun} = -0.003$. This value was chosen carefully to maintain an equilibrium between the two objectives.

\subsubsection{Solving the Blind Spot Problem}

\begin{figure}[h]
    \centering
    \includegraphics[width=8cm]{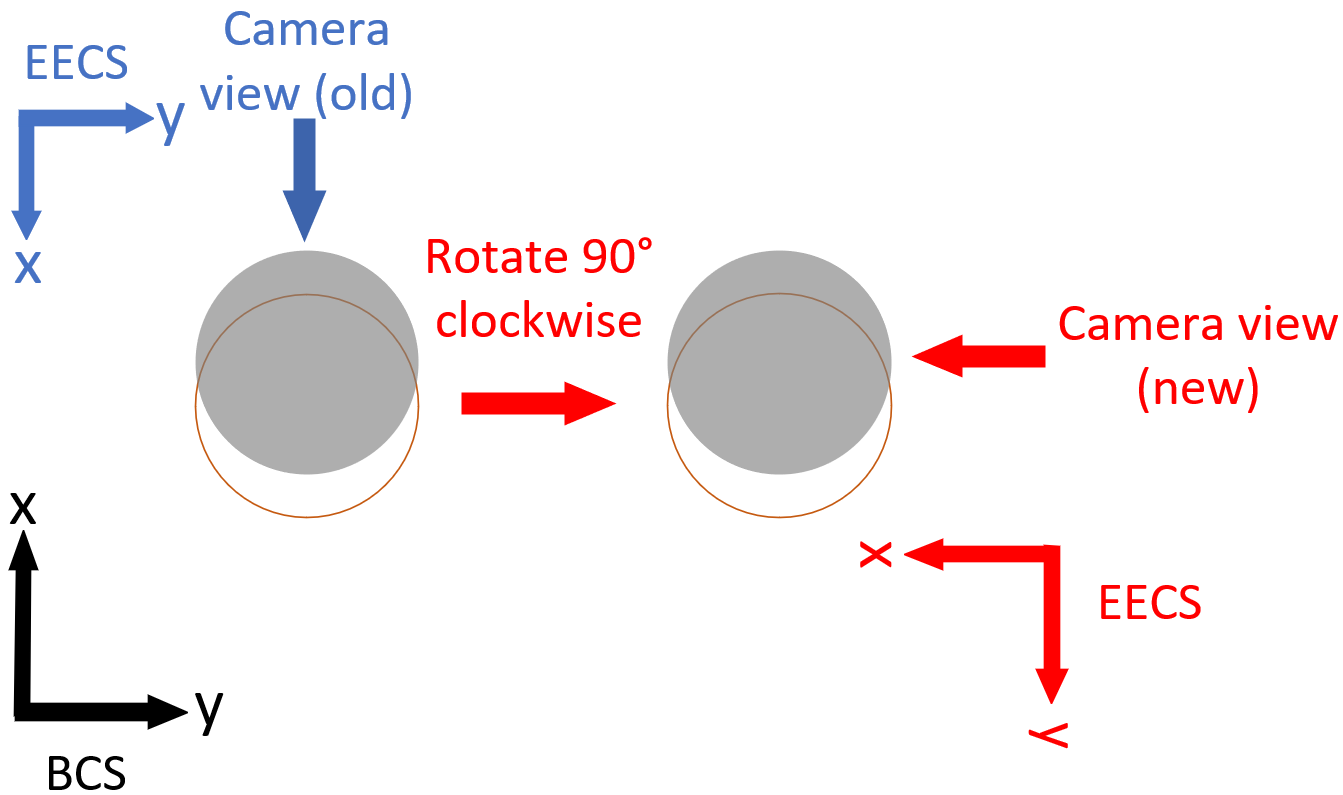}
    \caption{This is the top view of the camera's blind spot problem. It also contains the process of solving this problem. The grey translucent circles are the peg, and orange circles are the hole. The BCS indicates the base coordinate system, and the EECS indicates the end-effector coordinate system.}
    \label{fig:10}
\end{figure}

During the training process, to mitigate the risk of overfitting, we introduced a random selection mechanism for the initial pose of each episode, ensuring it was within 4mm of the ground truth position of the CPH. However, due to the usage of an eye-in-hand camera, there were instances where the peg obstructed the hole, creating a blind spot for the camera, as illustrated in the left portion of Figure \ref{fig:10}.

As depicted in Figure \ref{fig:10}, ideally, we would expect the agent to understand that it should move in the negative direction of the X-axis of the BCS in this particular scenario, just as a human would. However, this expectation is overly idealistic. A simpler yet effective solution is to rotate the final wrist of the robot arm by 90° clockwise, thereby eliminating the blind spot easily. Importantly, this rotation does not impact the network's performance. This is because all actions are performed within the EECS. In the case depicted in the left portion of Figure \ref{fig:10}, the network should make the movement decision in the positive direction of the X-axis of the original EECS. After the rotation, based on the image data, the hole is now on the right side of the peg. Therefore, according to the logical reasoning, the network should make the movement decision in the positive direction of the Y-axis of the new EECS. However, these two movements are exactly the same within the BCS. As a result, the rotation does not disrupt the network's decision-making logic.
\begin{figure}[!h]
    \centering
    \includegraphics[width=8cm]{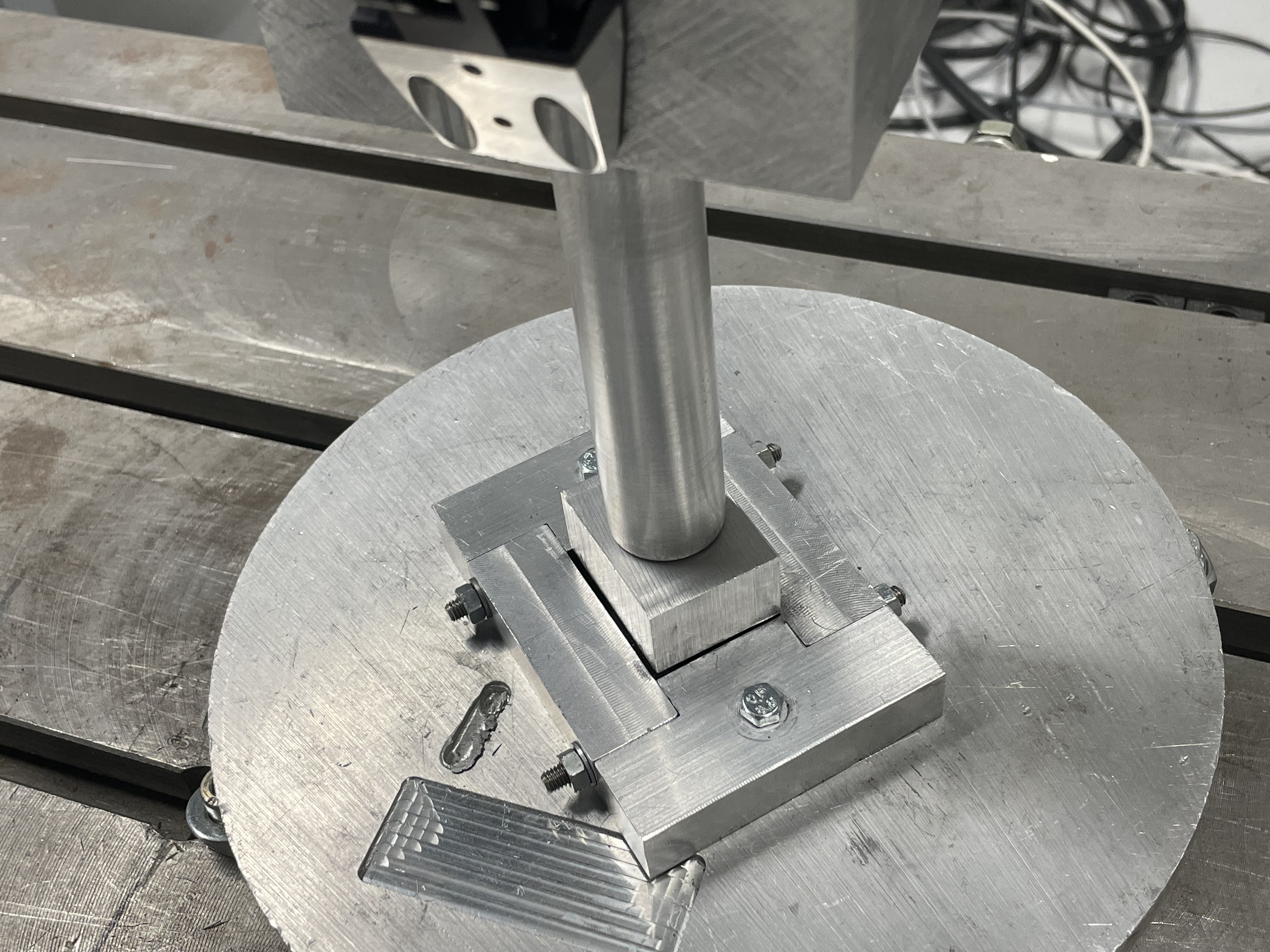}
    \caption{Rectangular peg-in-hole assembly.}
    \label{fig:94}
\end{figure}

Consequently, we implemented an Automatic Angle Selection Module (AASM). The AASM's main objective is to perform counterclockwise and clockwise rotations of the final wrist by 45 degrees, each done twice. Subsequently, the pose corresponding to the grayscale image with the highest number of black pixels is chosen as the initial pose.

\subsubsection{Assembly of Multiple Types of Pegs}
In order to ensure the versatility of the model, our training encompassed peg-in-hole assembly tasks involving various sizes and shapes. Specifically, in addition to the cylinder we used a metal peg with a rectangular cross-section, as shown in Figure \ref{fig:94}.

\subsection{Experiment}
During the experiment, we will first compare the performance of the well-trained agent on assembly tasks with different shapes, then we will compare performance on task finishing and violent contact avoidance of the agents with different training levels. Some performances are shown in the attached video (\url{https://youtu.be/GautzSyT1lE}).

\subsubsection{Performance of Agents with Different Training Levels on Assembly Task}
\begin{figure}[h]
\centering

\subfigure[]{
\includegraphics[width=5cm]{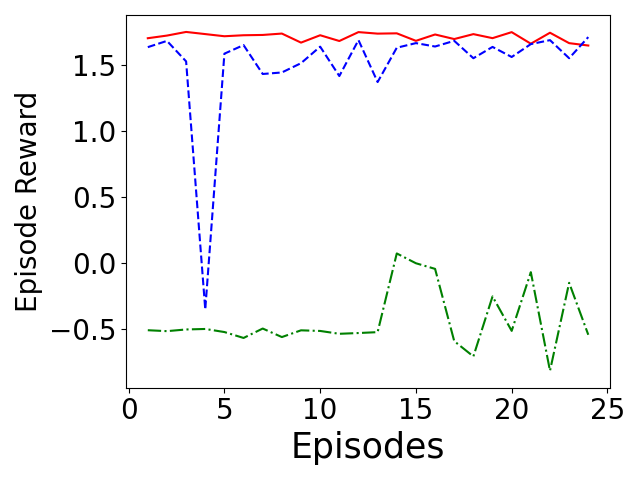}
}
\subfigure[]{

\includegraphics[width=5cm]{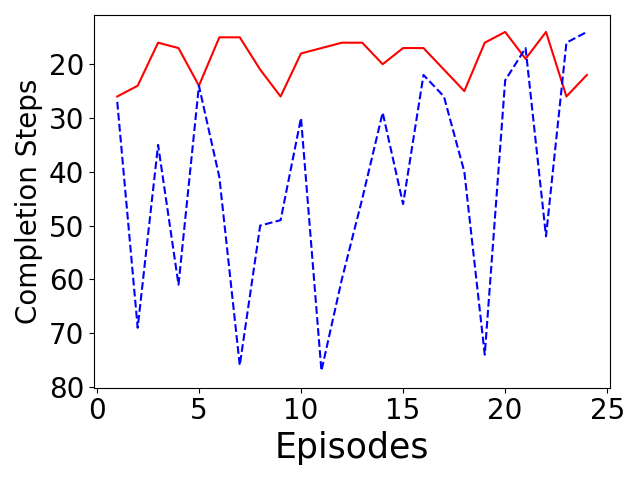}

}
\centering
\caption{The ER and completion steps comparisons. Red lines: fixed initial position with well-trained agent (FT). Blue lines: random initial position with well-trained agent (RT). Green lines:  fixed initial position with untrained agent (FNT).}
\label{fig:11}
\end{figure}

The performances are shown in Figure \ref{fig:11}, after organizing the data therein we obtain Table~\ref{table:3}. In Figure \ref{fig:11} (a), we compared the ER value of the FT, RT, and FNT. For the RT, there was one low ER value, which means that the agent failed to finish the assembly task in this episode. In the first 13 episodes, the ER of the FNT was quite stable and with low value; however, after the 13th episode, the ER values started to oscillate. This is because the network will only start updating the parameters by gradient descent if the number of agent's movement steps exceed a specific value. For those episodes with ER values under 0, the fewer steps the agent moves when it fails, the higher the ER value are. This could explain why the highest value of FNT is even higher than the minimum value of RT in Figure \ref{fig:11} (a). Meanwhile, for those episodes with ER values greater than 1, the fewer steps the agent moves before its success, the higher the ER value is.
Obviously, the well-trained agent's performance with the fixed initial position is best, which never failed and achieved the relative highest average ER value, which is $1.7120$. The well-trained agent failed once with the random initial position, and the average ER value was a litter bit lower. As for the agent without training, it failed in all the episodes.

\begin{table*}[h]
\caption{Comparison of performance on assembly task}\label{table:3}
\begin{tabular*}{\textwidth}{@{\extracolsep\fill}lcccc}
\bottomrule
     & Average ER &  \begin{tabular}[c]{@{}l@{}}No. of successes/ \\\ No. of attempts \end{tabular}      & Success rate & Average  completion steps\\ \midrule
FNT & -0.4325  &  0/24  &  0\%    &  - \\
RT & 1.5071   &  23/24  &  96\%    &  42 \\
FT & 1.7120   &  24/24  &  100\%    & 19 \\ \bottomrule
\end{tabular*}
\end{table*}

In Figure \ref{fig:11} (b), we compare the completion steps of the FT and RT.
It is important to know that we only counted the agent's completion step if it has successfully completed the assembly task, since the FNT could hardly finish the assembly task, and we do not have completion step statistics of the FNT in Figure \ref{fig:11} (b). The data shows that the FT has fewer and more stable completion steps then RT, which is consistent with Figure \ref{fig:11} (a). There is also a linear positive correlation between the completion steps and the completion time, with one step taking around one second.

From these analyses, we can draw the following conclusions: the well-trained agent could handle both the FT and RT situations well. Although there is once failure in RT situations, the agent could come out of the failure immediately and complete the following episode. Moreover, compared with the agent without training, the well-trained agent shows great advantages.

\subsubsection{Performance of Various of Peg-in-hole Assembly}
\begin{figure}[!h]
\centering

\subfigure[]{
\includegraphics[width=5cm]{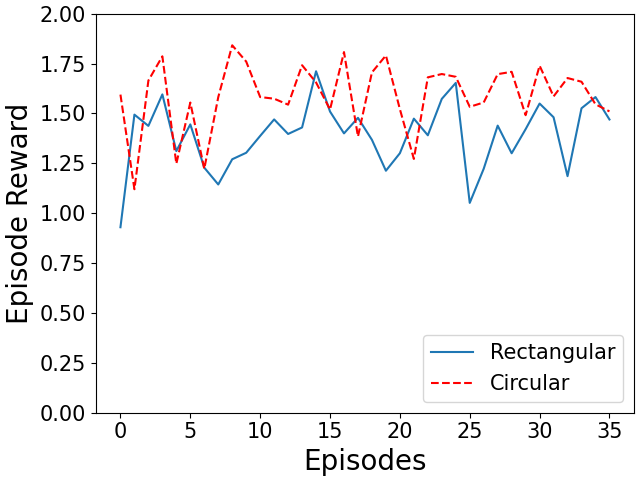}
}
\subfigure[]{

\includegraphics[width=5cm]{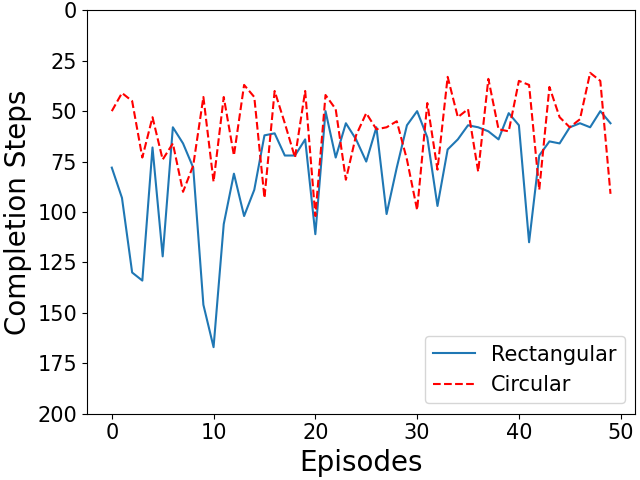}

}
\centering
\caption{Comparison of the performance of various peg-in-hole assembly tasks}
\label{fig:13}
\end{figure}

 In this subsection, we compare the performances of assembling a circular peg with that of a rectangular one in the assembly task, as depicted in Figure \ref{fig:13}. It should be noted that, when it comes to rectangular peg-in-hole assembly, if the initial relative positions of the holes are entirely random, the blind spot issue mentioned earlier may still persist, and rotating the peg angle alone cannot address this problem. As a result, when dealing with the rectangular peg-in-hole assembly task, the relative positions between the peg and hole are not entirely randomized. Once we have quantified the data from the Figure \ref{fig:13}, we can generate Table \ref{tab:44_0}.

\begin{table}[h]
\caption{Performance of different shapes of peg-in-hole assembly tasks}\label{tab:44_0}

\begin{tabular}{lcc}
\bottomrule
     & \textbf{ER} &  \textbf{Completion steps}     \\ \midrule
Circular peg & 1.5899  &  58.86  \\
Rectangular peg & 1.3927  &  77.16  \\\bottomrule
\end{tabular}

\end{table}

Based on Figure \ref{fig:13} and Table \ref{tab:44_0}, it is evident that the episode reward and the number of completion steps required for the circular assembly task outperform those of the rectangular task. This outcome is logical because in the rectangular assembly task, the robot arm needs to consider or perform an additional rotation operation in the $z$-axis direction, which negatively impacts both the correct behavioral choices of the robot arm and the execution efficiency. It should be noted that the initial relative positions between the peg and hole were not entirely random but rather somewhat constrained to avoid blind spots that cannot be resolved in the rectangular assembly.

\subsubsection{Performance of Agents with Different Training Levels on Avoiding Violent Collision or Contact}
\begin{figure}[!h]
    \centering
    \includegraphics[width=10cm]{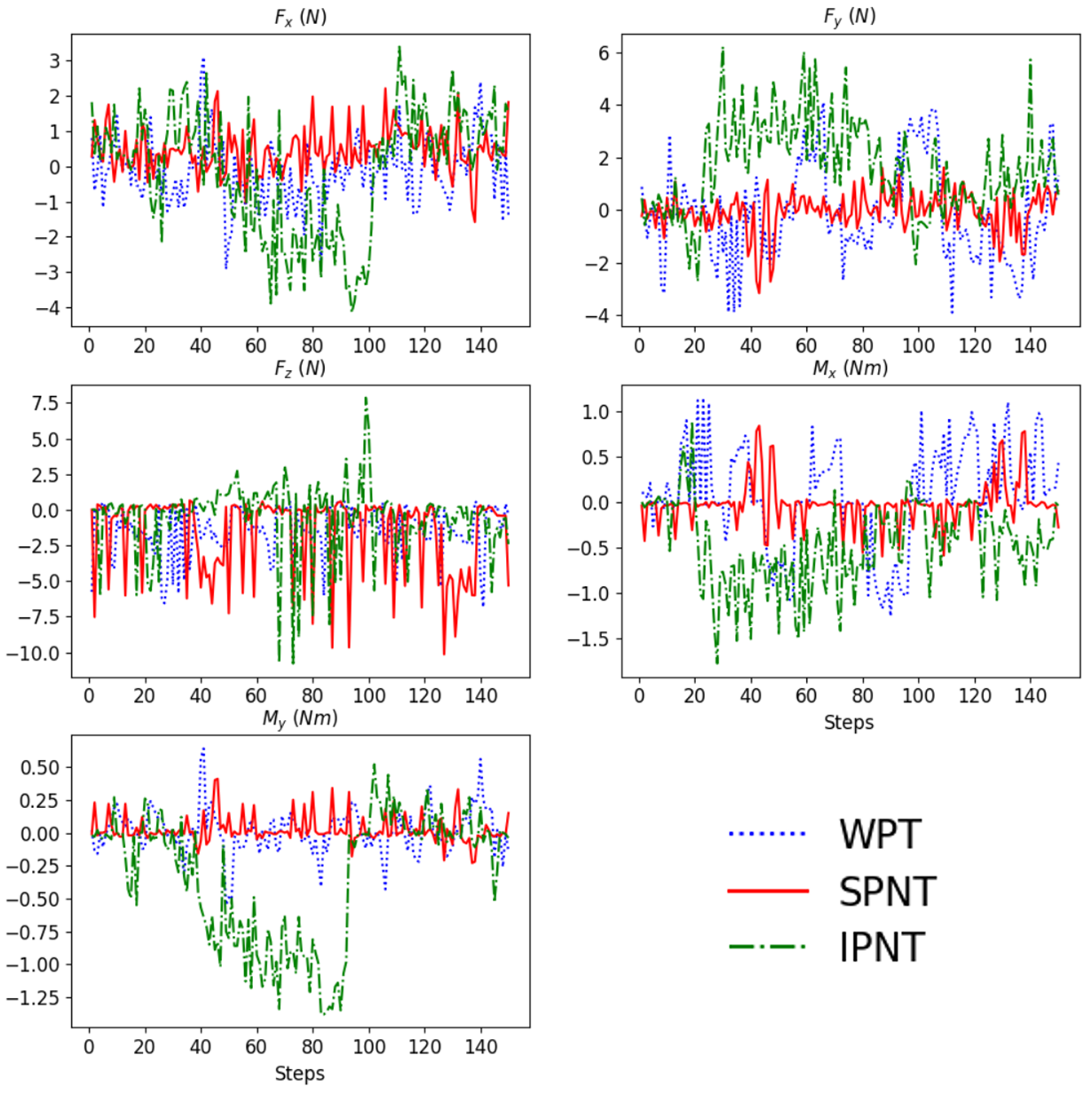}
    \caption{The data from the F/T sensor. Blue lines: the whole assembly process of the agent with training (WPT). Red lines: the searching process of the agent without training (SPNT). Green lines: the insert process of the agent without training (IPNT).}
    \label{fig:12}
\end{figure}

To verify the agent's performance on avoiding violent collision or contact, the data from the F/T sensor during the assembly task is shown in Figure \ref{fig:12}.

In Figure \ref{fig:12}, we compare the performances of the agent with and without training; it should be noted that we separate the assembly process of the agent without training into two phases: the searching process—the agent places the peg center within the clearance region of the hole center—and the insert process—the agent pushes the peg to the desired position. The reason for this is that it is difficult for the less-well–trained agent to complete the searching process automatically and thus start the insert process. We need to align the peg center with the hole center manually so that the agent can start the insert process; meanwhile, the data shows a big difference between these two phases, so we think it fairer and more objective to separate and compare the performances in these two phases.

From Figure \ref{fig:12}, we could find that the SPNT data looks even better than the WPT data except for the $F_{z}$. This is because the agent could move freely and did not meet any obstacles except for the Z-axis downward movement during the searching process. However, the IPNT data shows the state of the agent is basically in the warning zone and often comes to the edge of a dangerous situation, and the $F_{z}$ value has come to $-10 N$ several times and the $M_{z}$ has even come to $-1.78 Nm$, which is already in a dangerous situation according to the standard of TCS. Moreover, there is every reason to believe that if the TCS had not intervened in time, the agent would have been put into dangerous situations more frequently. On the other hand, from the WPT data, we can see the agent is attempting to remain in the safety zone during the whole assembly task and rarely comes into the warning zone. 

Therefore, through these analyses, we can state that the agent would perform more wisely to avoid the violent collision or contact after being well trained. Meanwhile, the TCS also works perfectly.

\section{Conclusion}
In this paper, we propose an N-step Dueling-DQN method to fuse data from multiple sensors through a network to enable robotic arms to handle the peg-in-hole assembly task, in a similar way to human hand-eye collaboration. Even though the peg-in-hole task in this paper is full of challenges—a rigid assembly task, high perception requirement, sensitivity to contact forces, randomness and uncertainty of initial relative position relationships between the peg and hole—the experiment, in reality, shows our method could successfully put an end to such kinds of challenging tasks and meet all the strict requirements; meanwhile, the TCS works well in preventing violent collision or contact, and the blind spot problem is also solved perfectly. All these designs make our method more general.

To ensure the feasibility of our approach in realistic environments, we carry out all our training and experiments in realistic environments, however it is still time-consuming.  In our future works, the development on more efficient leaning algorithms will be carried out.  And more sensitive force sensor and a new observation system will be installed to the experiment system to improve the performance of the system.

\section*{Declarations}

\subsection*{Funding}
This work has been carried out within the framework of the EUROfusion Consortium, funded by the European Union via the Euratom Research and Training Programme (Grant Agreement No. 101052200---EUROfusion). Views and opinions expressed are, however, those of the author(s) only and do not necessarily reflect those of the European Union or the European Commission. Neither the European Union nor the European Commission can be held responsible for them. This work was also supported by the Comprehensive Research Facility for Fusion Technology Program of China under Contract No. 2018-000052-73-01-001228, National Natural Science Foundation of China with Grant No.11905147 and the University Synergy Innovation Program of Anhui Province with Grant No. GXXT-2020-010.

\subsection*{Conflict of Interest}
The authors declare that they have no conflicts of interest.

\subsection*{Ethical Statement}
All user studies were conducted under LUT university guidelines and followed the protocol of the Responsible Conduct of Research and Procedures for Handling Allegations of Misconduct in Finland produced by the Finnish National Board on Research Integrity (TENK).

\subsection*{Author Contributions Statement}
Ruochen Yin wrote the  main manuscript text, Huapeng Wu helped with the construction of experimental platform, Ming Li helped with the writing, Yong Cheng helped with the data processing and Yuntao Song and Heikki Handroos edited the manuscript. All authors reviewed the manuscript.

\bibliography{sn-bibliography}

\end{document}